\documentclass[runningheads]{llncs}

\usepackage{eccv}

\usepackage{eccvabbrv}

\usepackage{graphicx}
\usepackage{booktabs}

\usepackage[accsupp]{axessibility}  

\usepackage{hyperref}

\usepackage{orcidlink}

\usepackage{multicol}
\usepackage{multirow}
\usepackage{wrapfig}
\usepackage[export]{adjustbox}

\usepackage[dvipsnames]{xcolor}
\usepackage{pgfplots}
\pgfplotsset{width=10cm,compat=1.9}
\makeatletter
\newcommand{\@chapapp}{\relax}%
\makeatother
\usepackage[title,toc,titletoc,header]{appendix}
\usepackage[ruled,vlined]{algorithm2e}

\begin{document}
\newcommand{\image}{\mathbf{x}}
\newcommand{\latent}{\mathbf{z}}
\newcommand{\camera}{\mathbf{c}}
\newcommand{\genset}{\mathcal{V}}
\newcommand{\condset}{\mathcal{C}}
\newcommand{\params}{\theta}

\newcommand{\score}{\mathbf{s}_\params}
\newcommand{\webpage}{\url{https://yorkucvil.github.io/PolyOculus-NVS/}}
\newcommand{\suppwebpage}{\href{https://yorkucvil.github.io/PolyOculus-NVS/}{Project Webpage}}

\newcommand{\dir}{\mathbf{d}}
\newcommand{\ray}{\mathbf{r}}
\newcommand{\rays}{\mathcal{R}}
\newcommand{\point}{\mathbf{p}}
\newcommand{\fundamental}{\mathbf{F}}

\newcommand{\refeq}[1]{Eq.\ \ref{#1}}
\newcommand{\reffig}[1]{Fig.\ \ref{#1}}
\newcommand{\reftab}[1]{Tab.\ \ref{#1}}
\newcommand{\refsec}[1]{Sec.\ \ref{#1}}
\newcommand{\refalgo}[1]{Algorithm\ \ref{#1}}

\newcommand{\rulesep}{\unskip\ \vrule\ }

\newcommand{\plotxtick}{0,20,40,60,80,100}
\newcommand{\plotytick}{0,20,40,60,80,100}
\newcommand{\legendstyle}{nodes={scale=0.5, transform shape}}
\newcommand{\plotheight}{4cm}
\newcommand{\widthscale}{1.2}
\newcommand{\marksize}{0.8pt}

\newcommand{\comment}[1]{\textcolor{blue}{\bf #1}}
\newcommand{\jason}[1]{\textcolor{blue}{\bf Jason: #1}}
\newcommand{\fereshteh}[1]{\textcolor{violet}{Fereshteh: #1}}
\newcommand{\kosta}[1]{\textcolor{blue}{\bf #1}}
\newcommand{\marcus}[1]{\textcolor{orange}{Marcus: #1}}
\newcommand{\ttaa}[1]{\textcolor{teal}{Tristan: #1}}


\newcommand{\checkme}[1]{#1}

\title{PolyOculus: Simultaneous Multi-view Image-based Novel View Synthesis} 

\titlerunning{PolyOculus: Simultaneous Multi-view Image-based NVS}

\author{
{Jason J. Yu\inst{1,2}\orcidlink{0009-0008-0909-4249}} \and
Tristan Aumentado-Armstrong\inst{1,2}\orcidlink{0000-0003-4876-7246} \and
Fereshteh Forghani\inst{1} \and
Konstantinos G. Derpanis\inst{1,2,3} \and
Marcus A. Brubaker\inst{1,2}\orcidlink{0000-0002-7892-9026}
}

\authorrunning{J.J.~Yu et al.}

\institute{
York University \and Vector Institute for AI
\and Samsung AI Centre Toronto
}

\maketitle

\begin{abstract}
This paper considers the problem of generative novel view synthesis (GNVS), generating novel, plausible views of a scene given a limited number of known views. 
Here, we propose a set-based generative model that can simultaneously generate multiple, self-consistent new views, conditioned on any number of views.
Our approach is not limited to generating a single image at a time and can condition on a variable number of views.
As a result, when generating a large number of views, our method is not restricted to a low-order autoregressive generation approach and is better able to maintain generated image quality over large sets of images.
We evaluate our model on standard NVS datasets and show that it outperforms the state-of-the-art image-based GNVS baselines.
Further, we show that the model is capable of generating sets of views that have no natural sequential ordering, like loops and binocular trajectories, and significantly outperforms other methods on such tasks.
Our project page is available at: \url{https://yorkucvil.github.io/PolyOculus-NVS/}.
\keywords{Novel view synthesis \and Generative \and Diffusion \and Multi-view}
\end{abstract}
\section{Introduction}
The goal of novel view synthesis (NVS) is to predict scene appearance from new perspectives, given multi-view posed images.
\textit{Generative} NVS (GNVS) considers the case where NVS is severely under-constrained:
for instance, given a single image of a hallway with an unobserved room, can we generate plausible novel views from \textit{inside} the room?
This is a conditional generative modeling problem, asking for images that match the previously observed views with potentially limited content overlap, and their implied 3D structure.
Existing work is either limited in its extrapolative capacity (e.g., cannot move the camera arbitrarily) or suffers from cross-view inconsistency (i.e., implausible changes between generated frames).
In this work, we mitigate the latter problem without incurring the former limitation by addressing a significant issue with the ordered autoregressive approach that characterizes existing image-based GNVS models.
Our primary innovation is a reformulation of the model as \textit{set-to-set} generation, meaning we condition on a set of posed images to generate another set of output views in a simultaneous and self-consistent manner. 
The resulting model outperforms existing methods in terms of image quality but also removes a fundamental yet previously unquestioned assumption about the ordered nature of viewpoint trajectories.
By doing so, our set-based method particularly improves performance on trajectories where a natural ordering is unclear. 

\begin{figure}[t]
    \centering
    \includegraphics[width=1.0\linewidth]{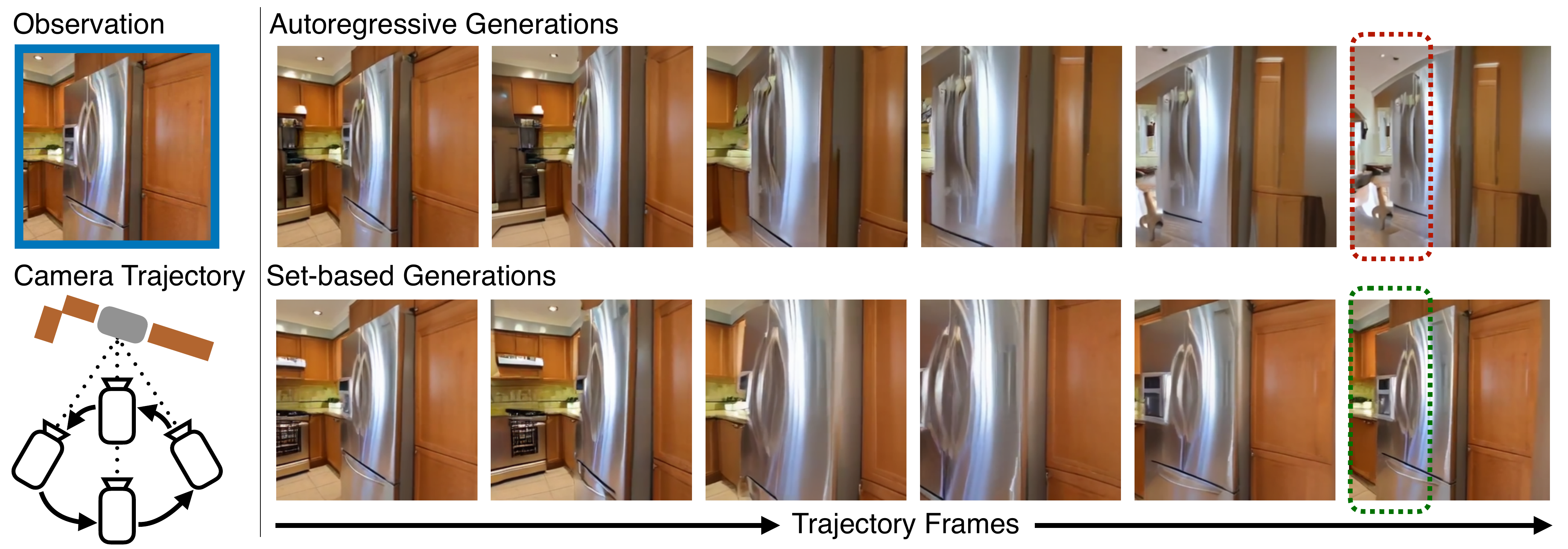}
    \caption{
        A ``loop inconsistency'', in which an image-based GNVS network fails to retain consistency with previous generations, and its resolution with a set-based approach.
        \textbf{Left:} An observed (real) image from which the synthesized views are conditioned, and camera trajectory (a ``spinning'' motion pointed at a fixed target) which temporarily conceals parts of the scene.
        \textbf{Top row:} An autoregressive strategy `forgets' its prior outputs and invents inconsistent content (e.g., compare the marked areas in the final column to the observation).
        \textbf{Bottom row:} Our method constructs self-consistent views by
        (i) more intelligent conditioning on the most relevant images
        and 
        (ii) simultaneous generation of a multi-view set, allowing mutual constraints within the sampling process.
    }
    \label{fig:fridge}
\end{figure}

NVS scenarios form a continuum between the well-determined case, and the GNVS case, which is largely decided by the number of available observations and their relation to the desired views (i.e., proximity in camera parameter space).
In the well-determined case, scene content has significant coverage from disparate viewpoints (relative to the target view distribution), which strictly constrains the 3D scene structure and minimizes ambiguity.
In contrast, the GNVS case, our focus, is less constrained when as few as one observation is given, or the model is queried for a viewpoint far from any of the given views.
Hence, generative models are a natural choice for representing this uncertainty. 

As conditional generative models increase in quality and scope of applicability (e.g., \cite{po2023state,croitoru2023diffusion}), interest in GNVS has surged as well, utilizing various forms and levels of inductive biases.
Rendering-based methods \cite{szymanowicz23viewset_diffusion,anciukevicius2022renderdiffusion,tewari2023diffusion,chan2023genvs} sample a 3D representation of the scene, which is then rendered to novel views. 
Such views are 3D consistent by design but can be limited by the spatial bounds of the representation.
In contrast, image-based models \cite{rombach2021geometry,ren2022look,watson2022novel,liu2023zero1to3,poseguideddiffusion,Yu2023PhotoconsistentNVS} approach NVS as an image-to-image problem without strong 3D priors.
These methods are not spatially bounded but necessitate learning 3D consistency from multi-view data.
Such approaches typically use (first-order) autoregressive sampling to generate longer trajectories, where images are generated one at a time, conditioned on only a single other view.
However, this sampling method
(i) accumulates errors over longer sequences, 
(ii) creates inconsistencies between near-viewpoint generations, due to divergent conditioning histories (e.g., when generating a sequence that loops back to its start), and
(iii) requires an \textit{ordering} of the views, which may not naturally exist (e.g., with loops and polyocular trajectories, as shown later). 
\reffig{fig:fridge} illustrates these issues, particularly (i) and (ii).

In this paper, we introduce a novel \textit{set-to-set} diffusion model for GNVS that both conditions on and generates view-sets of arbitrary cardinality in a permutation-invariant manner.
While it theoretically supports operating over arbitrarily sized sets, practical constraints (e.g., memory, run-time, and generalization) exist, which we circumvent with a novel hybrid sampling strategy.
Together, our model and sampling algorithm improves on the previously mentioned drawbacks of image-based GNVS methods.
Empirically, our contributions greatly improve image quality and realism, while specifically enhancing GNVS of view-sets \textit{without a natural order}, compared to an arbitrarily imposed ordering.

\section{Related Works}

\noindent
\textbf{Novel View Synthesis (NVS).}
Constructing new views of scenes from multi-view images has
a long history in computer vision and graphics (e.g., \cite{seitz1995physically,scharstein1996stereo,chen1993view,laveau19943,avidan1997novel}).
Such techniques roughly span two axes: image-based vs.\ geometry-based rendering (how much geometry is necessary \cite{shum2000review,chan2007image}), and interpolative vs.\ generative synthesis (how much \textit{extrapolation} is used).
Neural Radiance Fields (NeRFs), a recent interpolative method, renders views from a 3D geometry field  (e.g., \cite{mildenhall2021nerf,barron2023zip,xie2022neural,gao2022nerf}); however, it not only relies on per-scene optimization, but it is also unable to generate realistic details in unseen areas.
Our work, in contrast, eschews inferring geometry, as it entails solving an ill-posed problem that may not necessarily be required for realism or consistency. 
Instead, we focus on the \textit{generative} case, where the model is tasked to indefinitely extend the scene from minimal initial information (e.g., a single image).
Existing works that directly generate 3D structures often have limited scene extents, prohibiting indefinite spatial exploration (e.g., \cite{kim2023neuralfield,bautista2022gaudi}).
More recent methods extrapolate by combining autoregressive generation with 3D structure.
DFM~\cite{tewari2023diffusion} uses a generalizable NeRF \cite{yu2021pixelnerf} with a diffusion model, while GeNVS \cite{chan2023genvs} conditions a diffusion model on 3D features. 
The computational cost of sampling is generally very high for these methods, 
compared to our approach (e.g., limiting resolution; see Sec.~\ref{sec:seq_experiments}).

Our work is most similar to \textit{image-based} GNVS methods, generally based on pose-aware conditional diffusion models. 
A number of approaches (e.g., \cite{liu2023zero1to3,deitke2023objaverse,watson2022novel,szymanowicz23viewset_diffusion,anciukevicius2022renderdiffusion})
focus on object-centric NVS, whereas our method operates on full scenes and enables indefinite extrapolation.
In the scene-centric domain, GeoGPT \cite{rombach2021geometry}, LookOut \cite{ren2022look}, PhotoCon \cite{Yu2023PhotoconsistentNVS}, and Tseng et al.\ \cite{poseguideddiffusion} generate autoregressive sequences; however, while scene structure is coarsely preserved, dramatic changes can still occur across frames.
Previous video generative models struggle with similar challenges (e.g., \cite{geyer2023tokenflow,blattmann2023align}), albeit without the {3D} aspects. 
These methods suffer from error accumulation across trajectories and ``loop inconsistencies'',  where returning to the same viewpoint does not preserve  previously observed structure.
We show that a set-based approach can substantially mitigate such issues.

\noindent
\textbf{Set Learning.}
Set-valued data are common in computer vision
(e.g., point cloud processing 
     \cite{qi2017pointnet,qi2017pointnet++,aoki2019pointnetlk}, 
    anomaly detection 
    \cite{ravanbakhsh2016deep,lee2019set}, 
    and feature matching 
    \cite{sarlin2020superglue})
necessitating 
in/equivariance to permutation and cardinality \cite{zaheer2017deep,segol2019universal,zhang2021multiset}.
Similarly, many problems are naturally formulated 
as set prediction tasks,
including
    detection \cite{carion2020end,he2022voxel}, 
    and shape generation \cite{kim2021setvae,luo2021diffusion}.
For NVS, while prior work considered set-based encodings \cite{sajjadi2022scene}, it did not examine set-structured generation.

In this work, we devise a \textit{set-to-set} model by generating a set of NVs, given a set of conditioning images.
Unlike autoregression, this avoids 
an arbitrary ordering, 
which can cause difficulties 
\cite{zhang2019fspool,zhang2019deep}.
For instance, \textit{polyocular sequences} 
(e.g., binocular, to mimic humans or higher-order, for the multi-camera setups in autonomous driving \cite{yogamani2019woodscape} and elsewhere \cite{flynn2019deepview}) have no single natural ordering.
Finally, autoregression 
(e.g., \cite{ren2022look,rombach2021geometry,Yu2023PhotoconsistentNVS,liu2021infinite})
often accumulates errors over trajectories or incurs inconsistencies at overlapping views arrived at via different paths.
In contrast, by generating \textit{sets} of frames, we can mitigate such divergences.

\section{Technical Approach}
In this section, we first briefly review relevant background on diffusion models, then describe our set-based multi-view diffusion model, and finally detail methods for practically sampling large novel view-sets using our model.
\\\\
\noindent\textbf{Background.}\ Diffusion models \cite{sohl2015deep,song2020denoising,ho2020denoising} 
are generative models formulated as the reverse of a forward noising process.
This process is discretized into $t \in \{0,\ldots,T\}$ steps, where the value of the process $\image_T$ is approximately normal, and
$\image_0$ follows the data distritution.
Sampling the reverse process is often performed by iteratively applying a reparameterized noise estimator with learned parameters $\theta$ with hyper-parameters $\beta_t$, $\alpha_t = 1-\beta_t$, and $\Bar{\alpha}_t = \prod_{s=1}^t \alpha_s$:
\begin{align}
    \image_{t-1} = 
    \frac{1}{\sqrt{\alpha_t}}\left(\image_t - \frac{1-\alpha_t}{\sqrt{1-\Bar{\alpha}_t}}\epsilon_\theta(\image_t,t)\right) + 
    \beta_t \zeta_t ,
    \label{eq:reverse_reparam}
\end{align}
where ${\zeta}_t \sim \mathcal{N}(0,\mathbf{I})$.
Conditioning may be provided as additional inputs to $\epsilon_\theta$, which remain constant throughout the reverse process.

To reduce computational requirements, we use a latent diffusion model \cite{NEURIPS2021_701d8045,rombach2022high} that first maps the data into a latent space with reduced dimensionality, before the diffusion model is applied.
For our application, the autoencoder is fully convolutional to preserve the spatial structure of images.

\definecolor{ultramarine}{RGB}{0,32,96}
\definecolor{tagreen}{RGB}{102,204,0}
\definecolor{tagreent}{RGB}{92,194,0}
\definecolor{tadblue}{RGB}{1,11,154}
\definecolor{talblue}{RGB}{82, 139, 224}
\definecolor{taorange}{RGB}{250,128,4}
\definecolor{taoranget}{RGB}{240,120,0}

\begin{figure}[t]
    \centering
    \includegraphics[width=0.99\linewidth]{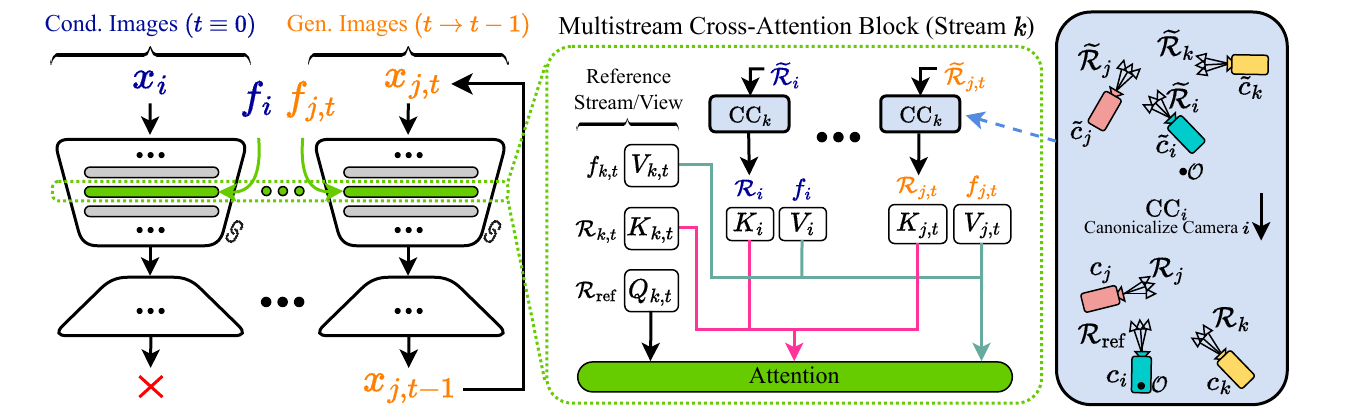}
    \caption{
    Overview of our set-based diffusion architecture for NVS. 
    Given a conditioning image set, 
    $\textcolor{tadblue}{ X_c =}$ $ \textcolor{tadblue}{ \{x_1,\ldots,x_{n_c}\} }$ \textcolor{tadblue}{(left inset, left-stream)}, 
    we 
    generate a set of novel views, $\textcolor{taoranget}{ X_{g,t} = \{ x_{n_c+1,t},\ldots,x_{n_c + n_g,t} \} }$ \textcolor{taoranget}{(left inset, right-stream)}. 
    Processing $\textcolor{tadblue}{ X_c }$ is identical to $\textcolor{taoranget}{ X_{g,t} }$,
        except time is fixed to $t\equiv 0$ (i.e., without noise).
    Simultaneous generation is performed by independently applying  
    the U-Net across streams,
    to each $ x\in \textcolor{tadblue}{ X_c }\cup \textcolor{taoranget}{ X_{g,t} } $, except at \textcolor{tagreent}{cross-attention (CA) layers (middle inset)}, which facilitate order-independent inter-stream dependencies.
    Each \textcolor{tagreent}{CA block} 
    combines  
    (i) same-layer features ($f_{j,t}$) and 
    (ii) camera information 
        ($\widetilde{c}_j$, with rays $\widetilde{\mathcal{R}}_j$)
        across streams. 
    For stream $i$,
        a \textcolor{talblue}{camera canonicalization block (right inset)}
        provides invariance to rigid transforms, via 
        $ {\mathcal{R}}_j = 
            \mathrm{CC}_i( \widetilde{\mathcal{R}}_j ) $,
        which treats $\widetilde{c}_i$ as a reference viewpoint.
    Then, with the reference rays as the queries, $Q_{i,t} = \mathcal{R}_\mathrm{ref}$,  
        attention is applied across all streams $j$,
        with the keys as transformed rays, $ \{ \mathcal{R}_j \}_j $,
        and the values as the features, $ \{ f_{j,t} \}_j $.
    }
    \label{fig:set_unet}
\end{figure}

\subsection{Set-based NVS Diffusion Model}
We begin by considering the distribution over sets of $N$ views, conditioned on their camera poses, 
$\camera = \{ \camera_1,\camera_2,\ldots,\camera_N \}$:
\begin{align}
    p_\params(\{ \latent_1, \latent_2,\ldots,\latent_N\} | \camera), \label{eq:uncond_prob}
\end{align}
where $\latent_n$ is the latent representation for view $n\in \{1,\ldots,N\}$.
Critically, this distribution should be permutation invariant since there may not be any natural ordering amongst the views.
Additionally, since the choice of global coordinates is arbitrary (i.e., only relative camera poses are meaningful),
the distribution should be invariant to global rigid transformations of the camera poses, $\camera$.

To construct such a \textit{set-based} diffusion model, 
we take inspiration from previous generative NVS methods \cite{watson2022novel,Yu2023PhotoconsistentNVS}, as well as video diffusion models \cite{ho2022video}, where separate streams of the model provide the noise estimate of each image.
Specifically, each view is processed by a U-Net \cite{ronneberger2015u} with shared weights, which only communicates between streams via attention layers, while the remaining layers operate on each view independently.
Since attention layers are permutation-equivariant by design, we can construct an ordering-independent noise estimator over sets of noisy latents, along with their camera poses:
\begin{align}
    \epsilon_\theta(\{(\latent_{1,t},\camera_1),\ldots,(\latent_{N,t},\camera_N)\},t). \label{eq:set_estimator}
\end{align}
Since reasoning about the scene across views relies deeply on knowledge of the (relative) camera geometry, 
the mechanism by which such information is injected into the generative process must be carefully designed.
In particular, camera poses are used to modulate the attention layers by injecting a camera ray representation into the keys and queries.
We adopt an image-based representation of 
camera rays, 
where per-pixel ray directions (determined by the camera parameters) are assembled into an array, followed by a fixed frequency-based Fourier encoding \cite{tancik2020fourier}
(e.g., see \cite{sajjadi2022scene,Yu2023PhotoconsistentNVS}; non-image-based NVS methods, such as NeRFs, utilize a similar representation as well).
However, since only the relative camera poses are meaningful, we \textit{canonicalize} the rays used by the attention block in each stream, such that the camera of the view processed by that stream is positioned at the origin with no rotation.
This enables mutually constrained generation (i.e., with each stream informed of each other) of arbitrary-sized sets in a permutation-invariant manner, 
while maintaining invariance to global rigid transforms.
The most similar method to ours, Yu et al.\ \cite{Yu2023PhotoconsistentNVS}, considered only the two-stream case, with one image acting as a fixed observation.
Our architecture, including the canonicalized rays, is illustrated in \reffig{fig:set_unet}.

While our model can sample without conditioning on scene information
via \refeq{eq:uncond_prob}, 
the primary cases of interest in GNVS involve conditioning on observed (or previously synthesized) frames.
Thus, our model (\refeq{eq:set_estimator}) can also be conditioned on observed views, using a small modification to the noise estimator:
\begin{align}
    \epsilon_\theta(\{(\latent_{1,t},\camera_1,t_1),\ldots,(\latent_{N,t},\camera_N,t_N)\}), \label{eq:set_estimator_cond}
\end{align}
where the time conditioning for all the views, $t$, is replaced with a separate time conditioning, $t_n$, for every view $\latent_n$.
Then, the model can flexibly treat inputs with $t_n=0$ as conditioning views (see also \cite{szymanowicz23viewset_diffusion}).
Such views are not noised; hence, they act as constant conditioning factors across the generative process.

\subsection{Sampling Large Sets of Views}\label{sec:samplinglargesets}
In principle, our model can operate over an arbitrarily large number of views.
Doing so in practice may be limited by the computational constraints caused by the quadratic time complexity of attention with respect to the number of views, and the model's ability to generalize to more views than available during training.
For these reasons, it may be desirable to limit the number of total views (the sum of conditioning and generated views) involved during sampling, and employ a semi-autoregressive approach to synthesize significantly more views than the imposed limit.
However, instead of generating a single view at a time, our model can generate and condition on sets of views.
By carefully selecting the sets and their order of generation, our model can mitigate image quality deterioration during autoregressive generation, and improve consistency for sets of views where there is no natural ordering.
In other words, for tractability, we can sample with a ``hybrid'' strategy, which uses autoregressive generation of sets of views, conditioned on sets of selected views.
Thus, selecting an appropriate conditioning hierarchy (i.e., which view-sets to output together and which to use for conditioning at each stage) is a new design space for our model.

\begin{figure}[t]
    \centering
    \includegraphics[width=\linewidth]{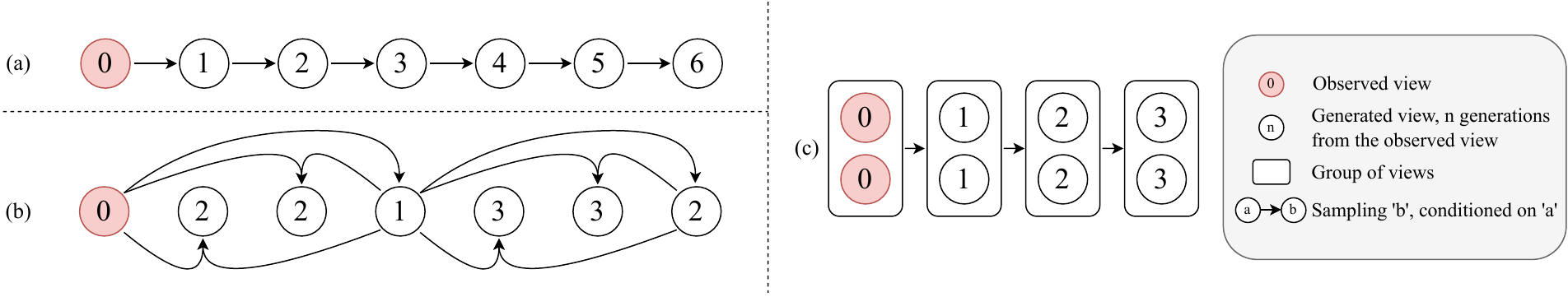}
    \caption{
    Illustration of different generation orders, and the sampling depth of each view from the observed image.
    The \textit{sampling depth} of each view from the observed view(s) is highlighted in \textcolor{red}{red} for
    (a) standard Markov autoregressive,
    (b) keyframed, and
    (c) grouped (\eg, stereo camera views) sampling.
    Notice that the view(s) with the largest sampling depth when sampling with (b) and (c) grows slower with respect to the total number of views, than with (a). 
    This reduces the error accumulation later views.
    }
    \label{fig:sampling_strats}
\end{figure}

Certain view-sets, such as those along a linear camera trajectory, have a natural ordering amenable to the standard Markov autoregressive sampling of prior work (e.g., \cite{Yu2023PhotoconsistentNVS,poseguideddiffusion,tewari2023diffusion,rombach2021geometry,ren2022look}).
Such methods can generate an arbitrary number of views by conditioning successive generations on previous ones.
Using a Markov assumption, the factorized generation distribution over views is:
\begin{align}
    p(\latent_1,\ldots,\latent_N) 
    = p(\latent_1) \prod_{n=2}^N p(\latent_n|\latent_{n-1}).
\end{align}
Hereafter,
we omit the dependence on the camera poses, $\camera$, for clarity.
A drawback of this method is the accumulation of errors over successive generations since views generated later in the sequence are conditioned on previous frames that may include errors themselves.
Further, by only considering the most recent frames, the model may \textit{forget} previously constructed scene content -- such errors lead to ``loop inconsistencies'' (see \reffig{fig:fridge}), in which frames generated for nearby cameras at different times do not match, due to different conditioning histories.

One way to reduce the error in the generated images is to reduce the \textit{sampling depth} 
of the views.
Consider a directed graph where nodes are images and directed edges represent conditional generations.
Then, the \textit{sampling depth} of a generated view is the minimum path length from the real initial frame.
A generation at high sampling depth, therefore, far from the original observation 
and, in image-based GNVS models, is liable to ``forget'' its constraints. 
The graph for first-order autoregressive generation is a chain; hence, its maximum sampling depth is proportional to the number of views.

In general, let $\genset_1,\dots,\genset_G$ be an ordered non-overlapping partition of all views to be generated and $\condset_i$ be the set of views to be conditioned on when generating $\genset_i$ such that $\condset_i \subseteq \cup_{j=1}^{i-1} \genset_j$, which ensures feasibility of the ordering.
Generation based on these sets can then be described as
\begin{equation}
p(\latent_1,\ldots,\latent_N) =
    \prod_{i=1}^{G} p(\{ \latent_j | \forall j \in \genset_i \} | \{ \latent_k |  \forall k \in \condset_i \} ).
\end{equation}
The maximum sampling depth of such a generation strategy is at most $G$. 
For instance, $\genset_1$ could generate a set of regularly spaced ``keyframes'', and subsequent $\genset_i$ would generate the in-between frames, conditioned only on neighbouring keyframes,
which could have a maximum sampling depth as low as two.

Similar strategies can be constructed in cases where there is no inherent or natural ordering of views, and the designation of keyframes may not be obvious.
For example, keyframe pairs can be considered for stereo pair trajectories.
Alternatively, when generating a ``cloud'' of views, the initial ``keyframes'' may be selected as evenly distributed among the set of views based upon their camera extrinsics, and ``in-between'' frames can be generated conditioned on their nearest, previously generated keyframes.
There are a multitude of generation strategies our model can support to manage the computational burden of generation while still minimizing sampling depth.
\reffig{fig:sampling_strats} illustrates three sampling strategies and the concept of sampling depth.
We explore different generation strategies in the experimental section, with further details in the supplement.

\section{Results}
\label{sec:setup_experiments}
\noindent\textbf{Experimental Setup.}\ We evaluate on the standard RealEstate10K \cite{zhou2018stereo} and Matterport3D \cite{chang2017matterport3d} datasets.
RealEstate10K consists of real-world multi-view image sets derived from videos.
The scenes provide rich, structured data with relatively large extents and camera motions,
    and have thus commonly been used by many prior works
    (e.g., \cite{Yu2023PhotoconsistentNVS,wang2021ibrnet,rombach2021geometry,ren2022look,tewari2023diffusion}).
Matterport3D contains 90 indoor building environments as textured meshes. 
Following previous work \cite{ren2022look,Yu2023PhotoconsistentNVS}, the meshes are converted into sets of views using a Habitat \cite{savva2019habitat} embodied agent.
In all cases, we use $256\times 256$ images derived from center crops.

Building on previous work \cite{Yu2023PhotoconsistentNVS}, our method is implemented as a latent diffusion model \cite{rombach2022high}, based on a VQ-GAN autoencoder \cite{esser2021taming}.
During training, for a given scene, we uniformly select $n\sim\mathcal{U}([1,\ldots,5])$ total views, then set $ n_c \sim \mathcal{U}([0,\ldots,n-1])$ of them to act as conditioning images (which have fixed diffusion time embedding, $t\equiv 0$; see \reffig{fig:set_unet}).
Our choice of the total number of views used during training is determined by available computational resources.
We utilize the standard DDPM noising process \cite{ho2020denoising}, with a modified denoising score matching loss that masks out the conditioning views.
Training and testing is performed using 16 and 1 RTX6000 GPUs with 24gb of memory, respectively.

We evaluate using the standard experimental settings of prior work (e.g., \cite{ren2022look,rombach2021geometry,Yu2023PhotoconsistentNVS}),
    while comparing to the state-of-the-art baselines PhotoCon \cite{Yu2023PhotoconsistentNVS}, and DFM \cite{tewari2023diffusion}.
PhotoCon uses a similar image-to-image diffusion approach to our method but only generates single novel views from single observations.
We also include two other image-to-image methods: GeoGPT \cite{rombach2021geometry} and Lookout \cite{ren2022look}.
DFM uses a diffusion method that samples NeRFs, conditioned on the observations, from which novel views are obtained by standard NeRF rendering \cite{mildenhall2021nerf}.

Because our method is trained in a set-based manner, the flexibility of our model enables a variety of \textit{sampling strategies}.
In particular, we compare three different strategies.
First, we consider 
the same method as PhotoCon and many previous GNVS methods: autoregressive sampling with a first-order Markov assumption (Ours-Markov).
Second, we naively sample all novel views simultaneously (Ours-1step).
Finally, we consider \textit{keyframe sampling} (Ours-KF), which was motivated in \refsec{sec:samplinglargesets}.
This strategy first generates sets of keyframes. 
Then, the remaining frames are conditioned on their neighboring keyframes, and are simultaneously generated with other views that share the same keyframes.

We use three metrics:
(i) when ground-truth sequences are available, we compute reference-based image quality via PSNR and LPIPS \cite{zhang2018unreasonable};
(ii) FID \cite{heusel2017gans}, which measures image quality or realism; 
and (iii) thresholded symmetric epipolar distance (TSED) \cite{Yu2023PhotoconsistentNVS}, which quantifies \textit{consistency} between views.
Briefly, TSED is computed over pairs of views with known camera poses; matched features between the images \cite{lowe1999object} are used to measure the error with respect to their expected epipolar geometry; image pairs with both sufficient matches (${\geq}T_\mathrm{matches}$) and low enough median error (${<}T_\mathrm{error}$) are counted as consistent, and the metric overall counts the proportion of consistent frames.
As in prior work \cite{Yu2023PhotoconsistentNVS}, we fix $T_\mathrm{matches} = 10$ and report TSED scores over a range of $T_\mathrm{error}$ thresholds.

\subsection{Generation on Sequential Views}
\label{sec:seq_experiments}
\begin{table}[tb]
    \footnotesize
    \centering
    \caption{
        RealEstate10K and Matterport3D reconstruction errors for short-term and long-term view extrapolation at $256\times256$.
        Bold numbers indicate best performance, while underlined numbers indicate the second best.
        Note that PSNR becomes less informative for distant frames, and does not necessarily correlate with image quality.
    }
    \resizebox{0.9\linewidth}{!}{
        \begin{tabular}{c|cc|cc||cc|cc}
            \multirow{3}{*}{Method} & \multicolumn{4}{c||}{RealEstate10K} & \multicolumn{4}{c}{Matterport3D}\\
             & \multicolumn{2}{c}{Short-term} & \multicolumn{2}{c||}{Long-term} & \multicolumn{2}{c}{Short-term} & \multicolumn{2}{c}{Long-term} \\
             & LPIPS $\downarrow$ & PSNR $\uparrow$ & LPIPS $\downarrow$ & PSNR $\uparrow$  & LPIPS $\downarrow$ & PSNR $\uparrow$ & LPIPS $\downarrow$ & PSNR $\uparrow$\\
            \hline
            GeoGPT \cite{rombach2021geometry}       &           {0.444} &           {13.35} &           {0.674} &           { 9.54} &           {  -  } &           {  -  } &           {  -  } &           {  -  } \\
            Lookout \cite{ren2022look}              &           {0.378} &           {14.43} &           {0.658} &           {10.51} &           {0.604} &           {12.76} &           {0.739} &           {12.76} \\
            PhotoCon \cite{Yu2023PhotoconsistentNVS}&           {0.333} &           {15.51} & \underline{0.588} &           {11.54} &           {0.504} &           {14.83} &           {0.674} &           {13.00} \\
            Ours-Markov                             & \underline{0.314} &           {15.57} &           {0.613} &           {11.63} & \underline{0.483} &           {14.66} &           {0.671} &           {12.75} \\
            Ours-1step                              &           {0.326} &    \textbf{16.42} &           {0.635} &    \textbf{12.88} &           {0.512} &    \textbf{15.93} & \underline{0.626} &    \textbf{14.55} \\
            Ours-KF                                 &    \textbf{0.290} & \underline{16.10} &    \textbf{0.564} & \underline{11.94} &    \textbf{0.473} & \underline{15.06} &    \textbf{0.598} & \underline{13.56} \\
        \end{tabular}
    }
    \label{tab:seq_recon}
\end{table}

We first perform standard evaluations on RealEstate10K and Matterport3D using trajectories from the datasets.
Each method is provided a single observation using the first frame of each ground-truth trajectory.
The remaining novel views are synthesized by each method using the provided ground-truth camera poses.
Following previous work \cite{Yu2023PhotoconsistentNVS}, we select test trajectories with at least 201 available frames for Realestate10K, which are subsampled such that we generate 20 views per sequence.
For Matterport3D, we follow a similar test selection criteria but select sequences with at least 21 frames, without subsampling.

\begin{figure}[t]
    \centering
    \includegraphics[width=0.90\linewidth]{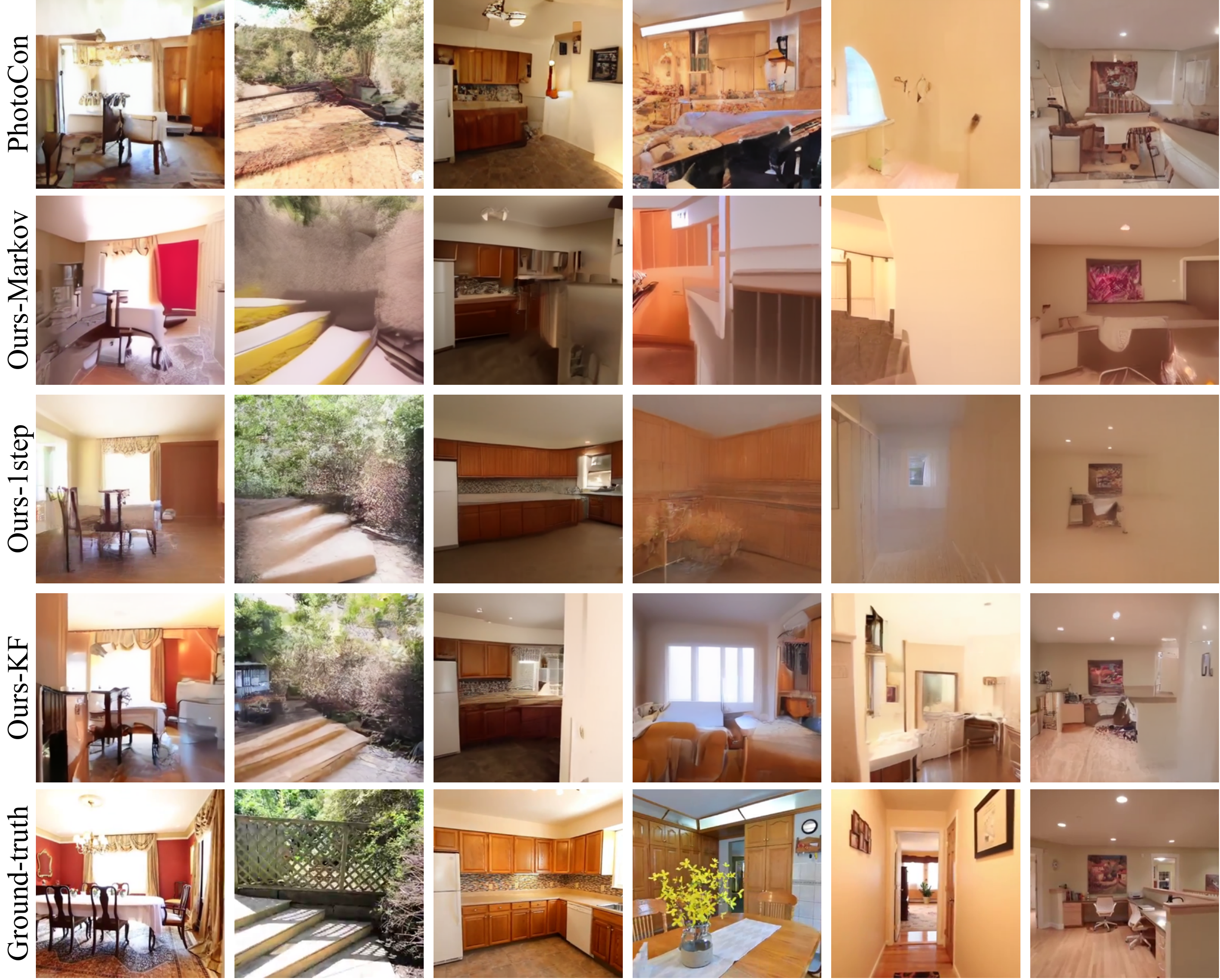}
    \caption{
    Visualization
    of the final frames of generated sequences on ground-truth trajectories.
    The image quality from our keyframed method  is higher than all other methods.
    }
    \label{fig:good_qual}
\end{figure}

We evaluate our method using the sampling strategy variations described in \refsec{sec:setup_experiments}.
For keyframed generation (Ours-KF), we 
generate keyframes along the sequence, each separated by two views.
A maximum of six total views are used while up to three of the views are used for simultaneous generation.
Results using different keyframing hyper-parameters are provided in the supplement.

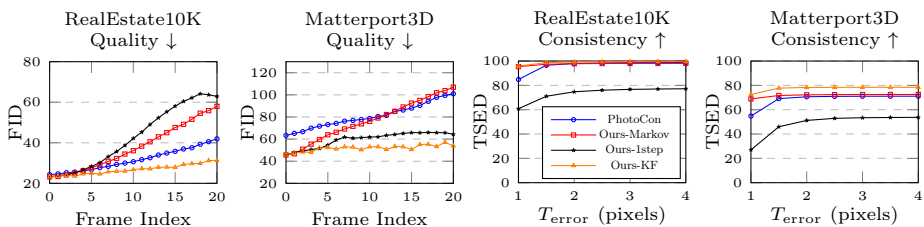
\begin{figure}[t]
    \centering
    \begin{subfigure}{0.24\linewidth}
        \pgfplotsset{width=1.3\linewidth,height=3.2cm,compat=1.18}
\begin{tikzpicture}
    \begin{axis}[
        title style={align=center, font=\scriptsize, yshift=-.5em},
        title={RealEstate10K\\Quality $\downarrow$},
        xlabel={Frame Index},
        ylabel={FID},
        xmin=0, xmax=20,
        ymin=20, ymax=80,
        xtick={0,5,10,15,20},
        ytick={0,20,40,60,80,100},
        legend pos=north west,
        legend style={nodes={scale=0.5, transform shape}},
        label style={font=\scriptsize},
        tick label style={font=\tiny},
        ymajorgrids=true,
        grid style=dashed,
        xlabel style={yshift=0.5ex},
        ylabel style={yshift=-1.5ex},
        mark size=0.8pt,
    ]
        \addplot[color=blue,mark=o,] coordinates {
            (0, 24.28090006699881)(1, 24.674042735910234)(2, 25.23376007268945)(3, 25.547281110854783)(4, 26.15348168802535)(5, 26.761569936294563)(6, 27.699912166963657)(7, 28.346736496348)(8, 29.305665914922486)(9, 29.976321882301022)(10, 30.65953359074956)(11, 31.647462664431032)(12, 32.61902125826293)(13, 33.95546167085729)(14, 34.91976868529588)(15, 35.77633529148716)(16, 36.933390346419515)(17, 37.555323108919424)(18, 39.267451981095974)(19, 40.54662671125158)(20, 41.95337459234463)
        };
        \addplot[color=red,mark=square,] coordinates {
            (0, 23.191309803640422)(1, 23.355105909798226)(2, 24.135172789174078)(3, 25.08736861597606)(4, 26.19127912060958)(5, 27.6740068813819)(6, 29.160867284164624)(7, 30.823110136240302)(8, 32.243681407223335)(9, 34.439615827914)(10, 36.08118425935129)(11, 38.339989564728114)(12, 40.48659628730235)(13, 42.93089553464233)(14, 44.98248592837746)(15, 47.537427714681314)(16, 49.11592524252933)(17, 51.58579397475148)(18, 54.57281497068726)(19, 55.832070512658106)(20, 57.97706690209509)
        };
        \addplot[color=black,mark=star,] coordinates {
            (0, 23.191309803640422)(1, 23.840016880791893)(2, 24.290369631594388)(3, 25.126952747132634)(4, 26.687537703473424)(5, 28.30704782992035)(6, 30.171070526957692)(7, 33.14657364875029)(8, 35.661597105282226)(9, 38.70878823143687)(10, 42.13716465630472)(11, 45.240843678994025)(12, 48.99712697382574)(13, 52.305656290992545)(14, 55.295088067459716)(15, 58.067667768479396)(16, 60.425504522545964)(17, 62.17573194514404)(18, 64.16765712406655)(19, 63.677485929467565)(20, 62.86161327893495)
        };
        \addplot[color=orange,mark=triangle,] coordinates {
            (0, 23.191309803640422)(1, 23.51614990681969)(2, 23.990226908808324)(3, 23.543027198182813)(4, 24.652106973566788)(5, 24.960601465382865)(6, 24.595232241333804)(7, 25.690130414383248)(8, 25.98251904596549)(9, 26.022328561151085)(10, 26.707767623294785)(11, 27.23416156086023)(12, 27.11967559177299)(13, 27.944219965430307)(14, 27.988278344121113)(15, 27.86203308760895)(16, 29.31409494242274)(17, 29.695871862108334)(18, 29.816069114379275)(19, 31.18597657793879)(20, 31.336071246278948)
        };
    \end{axis}
\end{tikzpicture}
    \end{subfigure}
    \hfill
    \begin{subfigure}{0.24\linewidth}
        \pgfplotsset{width=1.3\linewidth,height=3.2cm,compat=1.18}
\begin{tikzpicture}
    \begin{axis}[
        title style={align=center, font=\scriptsize, yshift=-.5em},
        title={Matterport3D\\Quality $\downarrow$},
        xlabel={Frame Index},
        ylabel={FID},
        xmin=0, xmax=20,
        ymin=20, ymax=130,
        xtick={0,5,10,15,20},
        ytick={0,20,40,60,80,100,120},
        legend pos=north west,
        legend style={nodes={scale=0.5, transform shape}},
        label style={font=\scriptsize},
        tick label style={font=\tiny},
        ymajorgrids=true,
        grid style=dashed,
        xlabel style={yshift=0.5ex},
        ylabel style={yshift=-2ex},
        mark size=0.8pt,
    ]
        \addplot[color=blue,mark=o,] coordinates {
            (0, 63.33417856993958)(1, 65.1714450809209)(2, 66.69597560831426)(3, 69.87103102383361)(4, 71.66961082735838)(5, 73.15756413981029)(6, 74.20162606967929)(7, 76.25126598652793)(8, 76.88152602480034)(9, 77.76863283210986)(10, 78.65626574996872)(11, 80.42938153879084)(12, 81.46700067027024)(13, 84.97817294910607)(14, 86.23248787681575)(15, 87.96358632944481)(16, 90.43659827349802)(17, 94.10948590587535)(18, 97.77266468985923)(19, 99.48077237427123)(20, 100.99214112104326)
        };
        \addplot[color=red,mark=square,] coordinates {
            (0, 45.665319490961735)(1, 46.71143776502515)(2, 50.82769974261987)(3, 55.47104619089029)(4, 59.59389964397292)(5, 63.86927632408839)(6, 66.13101137386693)(7, 69.13302475352071)(8, 71.51989166189105)(9, 73.7732714771933)(10, 76.06163749177125)(11, 79.07768732214794)(12, 82.2091662620818)(13, 85.12337869323198)(14, 89.03702626150093)(15, 92.54709875698134)(16, 95.0234133921663)(17, 98.37933475298453)(18, 101.88633130971238)(19, 103.8459062232032)(20, 106.95768260985491)
        };
        \addplot[color=black,mark=star,] coordinates {
            (0, 45.663996178020966)(1, 48.03145209316693)(2, 49.179725471143485)(3, 50.53753440367933)(4, 51.47900050575774)(5, 54.55537955631053)(6, 59.15557245407376)(7, 61.777676821661714)(8, 60.80778674024464)(9, 61.31241680202109)(10, 61.82557089371406)(11, 62.02435895158695)(12, 63.40170369423336)(13, 64.1592562130005)(14, 65.10581430940249)(15, 65.80863143114584)(16, 65.75201085069187)(17, 65.96295561203732)(18, 65.83827366809055)(19, 65.75915012855671)(20, 64.37026843649667)
        };
        \addplot[color=orange,mark=triangle,] coordinates {
            (0, 45.665319490961735)(1, 47.46340310107098)(2, 48.325166515348485)(3, 47.969777952972834)(4, 52.01942310889302)(5, 52.35791189883656)(6, 51.01787708126301)(7, 53.00134598933704)(8, 52.7262525902961)(9, 50.33694402631414)(10, 53.004214076612186)(11, 53.077097445095035)(12, 50.66146087387486)(13, 53.25498313652031)(14, 53.04503362239478)(15, 50.606420302167095)(16, 55.08117340763101)(17, 55.38410363348578)(18, 53.22042679541269)(19, 57.068967823702934)(20, 53.893847304557994)
        };
    \end{axis}
\end{tikzpicture}
    \end{subfigure}
    \hfill
    \begin{subfigure}{0.24\linewidth}
        \pgfplotsset{width=1.3\linewidth,height=3.2cm,compat=1.18}
\begin{tikzpicture}
    \begin{axis}[
        title style={align=center, font=\scriptsize, yshift=-.5em},
        title={RealEstate10K\\Consistency $\uparrow$},
        xlabel={$T_\text{error}$ (pixels)},
        ylabel={TSED},
        xmin=1, xmax=4,
        ymin=0, ymax=100,
        xtick={1,2,3,4},
        ytick={0,20,40,60,80,100},
        legend pos=south east,
        legend style={\legendstyle},
        label style={font=\scriptsize},
        tick label style={font=\tiny},
        ymajorgrids=true,
        grid style=dashed,
        xlabel style={yshift=1ex},
        ylabel style={yshift=-2ex},
        mark size=\marksize,
    ]
        \addplot[color=blue,mark=o,] coordinates {
            (1.0,84.87241798298906)(1.5,96.52490886998785)(2.0,98.00729040097205)(2.5,98.45990279465371)(3.0,98.63912515188335)(3.5,98.72417982989064)(4.0,98.79404617253948)
        };
        \addplot[color=red,mark=square,] coordinates {
            (1.0,95.37970838396112)(1.5,97.38153098420413)(2.0,97.79769137302551)(2.5,97.9678007290401)(3.0,98.04374240583232)(3.5,98.06500607533414)(4.0,98.07411907654921)
        };
        \addplot[color=black,mark=star,] coordinates {
            (1.0,60.81409477521264)(1.5,71.05407047387607)(2.0,74.7296476306197)(2.5,76.03584447144594)(3.0,76.7071688942892)(3.5,77.01093560145809)(4.0,77.20838396111786)
        };
        \addplot[color=orange,mark=triangle,] coordinates {
            (1.0,95.90218712029161)(1.5,98.67253948967193)(2.0,99.27095990279466)(2.5,99.42891859052247)(3.0,99.52308626974484)(3.5,99.5595382746051)(4.0,99.56561360874849)
        };
        \legend{PhotoCon, Ours-Markov, Ours-1step, Ours-KF}
    \end{axis}
\end{tikzpicture}
    \end{subfigure}
    \hfill
    \begin{subfigure}{0.24\linewidth}
        \pgfplotsset{width=1.3\linewidth,height=3.2cm,compat=1.18}
\begin{tikzpicture}
    \begin{axis}[
        title style={align=center, font=\scriptsize, yshift=-.5em},
        title={Matterport3D\\Consistency $\uparrow$},
        xlabel={$T_\text{error}$ (pixels)},
        ylabel={TSED},
        xmin=1, xmax=4,
        ymin=0, ymax=100,
        xtick={1,2,3,4},
        ytick={0,20,40,60,80,100},
        legend pos=south east,
        legend style={\legendstyle},
        label style={font=\scriptsize},
        tick label style={font=\tiny},
        ymajorgrids=true,
        grid style=dashed,
        xlabel style={yshift=1ex},
        ylabel style={yshift=-2ex},
        mark size=\marksize,
    ]
        \addplot[color=blue,mark=o,] coordinates {
            (1.0,54.79206049149339)(1.5,69.09735349716446)(2.0,70.63327032136105)(2.5,70.96408317580341)(3.0,71.07277882797732)(3.5,71.10113421550095)(4.0,71.11058601134215)
        };
        \addplot[color=red,mark=square,] coordinates {
            (1.0,68.87523629489604)(1.5,71.67769376181474)(2.0,72.22589792060491)(2.5,72.39603024574669)(3.0,72.5)(3.5,72.51890359168243)(4.0,72.53780718336483)
        };
        \addplot[color=black,mark=star,] coordinates {
            (1.0,27.008506616257087)(1.5,45.97826086956521)(2.0,51.190926275992446)(2.5,52.89224952741021)(3.0,53.40264650283554)(3.5,53.596408317580334)(4.0,53.71455576559546)
        }; 
        \addplot[color=orange,mark=triangle,] coordinates {
            (1.0,72.48582230623819)(1.5,77.74574669187146)(2.0,78.28449905482042)(2.5,78.38846880907371)(3.0,78.44045368620039)(3.5,78.44990548204159)(4.0,78.44990548204159)
        };
    \end{axis}
\end{tikzpicture}
    \end{subfigure}
    \caption{
    Sample quality and consistency evaluated using FID and TSED, respectively, for RealEstate10K and Matterport3D.
    Evaluation is performed at $256\times256$ resolution.
    }
    \label{fig:seq_fid_tsed}
\end{figure}

With the known ground-truth frames, we perform standard reference-based evaluations in image space.
We consider the fifth and final generated images as short-term and long-term views, respectively.
As shown in \reftab{tab:seq_recon}, Ours-KF has the best performance in terms of perceptual similarity, and second best performance in terms of PSNR.
Ours-1step obtains the best PSNR; however, PSNR is known to not always correlate well with image quality.
Qualitative inspection on RealEstate10k, in \reffig{fig:good_qual}, shows that Ours-1step achieves better PSNR scores by generating images that are generally smooth with few fine details.
A similar qualitative result holds for Matterport3D, provided in the supplement.
Although reference-based evaluations are standard, they are less informative for extrapolated views far from the observation \cite{rombach2021geometry,Yu2023PhotoconsistentNVS}, which motivates the use of other metrics for evaluating GNVS.
Given that PhotoCon \cite{Yu2023PhotoconsistentNVS} consistently outperforms
GeoGPT \cite{rombach2021geometry} and Lookout \cite{ren2022look}, we focus on it as our primary baseline.

To evaluate the image quality as a function of distance from the observation, we evaluate FID between the generated views at each time-step and a fixed set of random test views.
Quantitative results in \reffig{fig:seq_fid_tsed} (left) show that Ours-KF better mitigates the deterioration of FID as trajectories grow, than the other methods, giving us the best image quality on both datasets.
The reduced degradation from keyframing can be observed qualitatively in \reffig{fig:good_qual}, for RealEstate10K.

FID measures independent image quality; low FIDs can be obtained by generating sets of high-quality images that are inconsistent as a whole.
Consistency evaluations using TSED on both datasets in \reffig{fig:seq_fid_tsed} (right), show Ours-KF does not sacrifice inter-frame consistency to obtain its improved image quality.
Ours-1step generally achieves lower consistency, as its over-smoothed generations reduce the number of feature matches.
Ours-KF and Ours-1step recover 148.8 and 60.0 average matches, respectively.
Thus, keyframing with our set-based model offers increased image quality while maintaining consistency.

The non-keyframed variants are two extreme strategies for sampling with our model.
Ours-Markov limits the number of views within a range seen during training but has the highest sampling depth possible, which causes a significant accumulation of errors.
Ours-1step has the lowest possible sampling depth, which mitigates autoregressive error accumulation, but suffers from generalization errors due to sampling significantly more views than seen in training. 
Our keyframing approach offers a middle ground, where both sources of performance degradation are mitigated, providing higher generation quality and consistency.

\noindent\textbf{Comparison to Rendering-based GNVS.}\ 
We also compare to a state-of-the-art rendering-based GNVS approach: DFM \cite{tewari2023diffusion}, an image-to-NeRF model that renders through an inferred geometry field.
Unlike image-based methods, which eschew geometry, DFM represents 3D geometry with a NeRF. 
Methods in this alternative paradigm require a very different sampling process: briefly, DFM first samples a NeRF \cite{yu2021pixelnerf} using a generative diffusion process defined over the rendered output images.
Views that participate in the NeRF sampling are \textit{target views}, whereas non-target views are obtained by simply rendering the NeRF.
While this potentially improves 3D consistency, it significantly increases the costs of both training and image generation.
This approach also complicates long-range extrapolation; while our method readily extrapolates from any starting image set, DFM requires costly target view generation to cover distant views. 
Ideally, all the synthesized views would be target views, but this is intractable, as computation is proportional to the number of targets.
As a result, DFM \cite{tewari2023diffusion} uses only a sparse set of target views.
For fairness, we limit the number of targets to a level tractable for evaluation and similar in overall computation time.  
Importantly, note that generating a DFM target view is not the same as NVS from our model, as it is optimized \textit{through} the NeRF. 
Further, we find that even the target views do not have the same image quality as our outputs.

Due to computational costs, DFM operates at a resolution of $128\times128$; thus, for fairness, we downsample our outputs to the same resolution. 
In our setup, sampling a 21-frame sequence from DFM using two target views takes ${\sim}29$ minutes and up to ${\sim}24$gb of memory, while our keyframed approach takes ${\sim}11$ minutes with ${\sim}5$gb of peak memory usage.
These computational considerations limit our DFM evaluation to use only up to two target views (DFM-2).
For a single target, we choose the last view, while for two targets, we choose the middle and last views.
This is analogous to keyframing with their target views.

\begin{figure}[t]
    \centering
    \begin{minipage}[b][][t]{.5\linewidth}
        \centering
        \captionof{table}{
            RealEstate10K reconstruction errors for short-term and long-term view extrapolation at $128\times128$.
        }
        \resizebox{\linewidth}{!}{
            \begin{tabular}{c|cc|cc}
                \multirow{2}{*}{Method} & \multicolumn{2}{c}{Short-term} & \multicolumn{2}{c}{Long-term} \\
                & LPIPS $\downarrow$ & PSNR $\uparrow$ & LPIPS $\downarrow$ & PSNR $\uparrow$\\
                \hline
                DFM-1 \cite{tewari2023diffusion}& \underline{0.217} &    \textbf{17.70} &    \textbf{0.469} & \underline{12.22} \\
                DFM-2 \cite{tewari2023diffusion}&           {0.255} & \underline{17.09} &           {0.484} &    \textbf{12.28} \\
                Ours-KF                         &    \textbf{0.210} &           {16.56} & \underline{0.471} &           {12.10} \\
            \end{tabular}
        }
        \label{tab:dfm_recon}
    \end{minipage}
    \begin{minipage}[b][][t]{.49\linewidth}
        \centering
        \begin{minipage}[b]{.48\linewidth}
            \pgfplotsset{width=1.3\linewidth,height=3cm,compat=1.18}
            \begin{tikzpicture}
                \begin{axis}[
                    title style={align=center, font=\scriptsize, yshift=-.5em},
                    title={RealEstate10K\\Quality $\downarrow$},
                    xlabel={Frame Index},
                    ylabel={FID},
                    xmin=0, xmax=20,
                    ymin=20, ymax=60,
                    xtick={0,5,10,15,20},
                    ytick={0,20,40,60,80,100},
                    legend pos=north west,
                    legend style={nodes={scale=0.5, transform shape}},
                    label style={font=\scriptsize},
                    tick label style={font=\tiny},
                    ymajorgrids=true,
                    grid style=dashed,
                    xlabel style={yshift=0.5ex},
                    ylabel style={yshift=-1.5ex},
                    mark size=0.8pt,
                ]
                    \addplot[color=blue,mark=pentagon,] coordinates {
                        (0, 22.575377178658414)(1, 23.23675138773649)(2, 25.05309571420139)(3, 27.228895242512692)(4, 30.812179278576878)(5, 34.38595403640255)(6, 37.92099313547402)(7, 41.77310101546465)(8, 44.806860993730595)(9, 47.829849041714056)(10, 50.199337915534045)(11, 52.05456852735034)(12, 52.143740454686736)(13, 51.21934124216358)(14, 49.62116627891089)(15, 47.52959530950369)(16, 44.39837285784233)(17, 40.97693269908797)(18, 37.30108829768898)(19, 34.242348141119095)(20, 32.16232196181437)
                    };
                    \addplot[color=black,mark=pentagon,] coordinates {
                        (0, 24.203520229116634)(1, 25.842598590453917)(2, 28.289119743817253)(3, 31.31884142040235)(4, 34.22192964931952)(5, 36.742465213364596)(6, 38.03759461885426)(7, 38.17640486681819)(8, 36.7790273957126)(9, 34.759572256120464)(10, 32.392956221407246)(11, 35.26632244078178)(12, 38.486501098667475)(13, 41.04009795916767)(14, 43.242811446577434)(15, 44.00061306099485)(16, 44.31714180299258)(17, 42.52261782166522)(18, 40.45842167774461)(19, 38.45615673792372)(20, 35.40208706456164)
                    };
                    \addplot[color=orange,mark=triangle,] coordinates {
                        (0, 21.47384410087284)(1, 21.775088908493046)(2, 22.229723666021812)(3, 21.935138305968223)(4, 22.886521520952954)(5, 23.248575452341242)(6, 22.813713460582335)(7, 23.593501882691044)(8, 23.929954387963022)(9, 23.764453006849863)(10, 24.488856822194123)(11, 24.942316881600732)(12, 24.509774458911124)(13, 25.47132718633884)(14, 25.584127599006877)(15, 25.115521302509507)(16, 26.26512821840643)(17, 26.566989396029555)(18, 26.629070505927473)(19, 27.783686142973465)(20, 27.71812381963673)
                    };
                \end{axis}
            \end{tikzpicture}   
        \end{minipage}
        \begin{minipage}[b]{.49\linewidth}
            \pgfplotsset{width=1.3\linewidth,height=3cm,compat=1.18}
            \begin{tikzpicture}
                \begin{axis}[
                    title style={align=center, font=\scriptsize, yshift=-.5em},
                    title={RealEstate10K\\Consistency $\uparrow$},
                    xlabel={$T_\text{error}$ (pixels)},
                    ylabel={TSED},
                    xmin=0, xmax=2,
                    ymin=40, ymax=100,
                    xtick={0.5,1,1.5,2},
                    ytick={0,20,40,60,80,100},
                    legend pos=south east,
                    legend style={\legendstyle},
                    label style={font=\scriptsize},
                    tick label style={font=\tiny},
                    ymajorgrids=true,
                    grid style=dashed,
                    xlabel style={yshift=1ex},
                    ylabel style={yshift=-2ex},
                    mark size=\marksize,
                ]
                    \addplot[color=blue,mark=pentagon,] coordinates {
                        (0.25,94.14034021871203)(0.5,97.10510328068042)(0.75,97.33900364520048)(1.0,97.37849331713244)(1.25,97.40886998784933)(1.5,97.41190765492101)(1.75,97.41190765492101)(2.0,97.41190765492101)
                    };
                    \addplot[color=black,mark=pentagon,] coordinates {
                        (0.25,94.85115431348724)(0.5,96.8955042527339)(0.75,97.001822600243)(1.0,97.01701093560146)(1.25,97.01701093560146)(1.5,97.02308626974482)(1.75,97.02612393681653)(2.0,97.02612393681653)
                    };
                    \addplot[color=orange,mark=triangle,] coordinates {
                        (0.25,57.85540704738761)(0.5,93.97630619684082)(0.75,97.76123936816525)(1.0,98.5419198055893)(1.25,98.75151883353584)(1.5,98.82746051032807)(1.75,98.86998784933171)(2.0,98.88517618469017)
                    };
                    \legend{DFM-1 ,DFM-2 ,Ours-KF}
                \end{axis}
            \end{tikzpicture}
        \end{minipage}
        \captionof{figure}{Comparisons with Ours-KF and DFM on FID and TSED at $128\times128$.}
        \label{fig:dfm_fid_tsed}
    \end{minipage}
\end{figure}

We compare Ours-KF to DFM using the same evaluations as PhotoCon on RealEstate10K but at a reduced resolution.
Reference-based evaluations in \reftab{tab:dfm_recon} show that our method is competitive with DFM, especially on long-term paths.
More importantly, the quantitative results in \reffig{fig:dfm_fid_tsed} show that \textit{our method produces higher quality images in terms of FID, despite sampling roughly three times faster at double the resolution}.
Unsurprisingly, DFM is able to obtain better consistency, as NeRFs have strong 3D constraints by design.
Overall, our method generates views with a consistency comparable to DFM but with superior image quality and lower computational requirements.

\subsection{Loop Closure}
Standard evaluations use ground-truth trajectories, which are often sequential in nature.
However, \textit{non}-sequential trajectories may be of interest, such as binocular vision.
One example, devised to test generalization, is the \textit{spin} trajectory introduced by Yu et al.~\cite{Yu2023PhotoconsistentNVS},
where the camera ends close to the initial observation.
When the trajectory is generated sequentially,
these spatially close cameras are distant in the number of sampling steps.
A Markov assumption over the view dependencies causes the model to ``forget'' about previous relevant views (i.e., the generation depth is too high), resulting in ``loop inconsistencies'' (e.g., \reffig{fig:fridge}).
For our keyframed approach, selecting keyframes that maximize the inter-camera distance of the initial frame-set increases coverage of the scene, allowing view dependencies to be captured without ordering (see \refsec{sec:samplinglargesets} and the supplement for details).
Considering all views at once achieves the same effect but requires the model to handle a larger set.
In this way, we can generate cyclical trajectories where the start and end frames should be consistent since frames near the end have low generation depth.

We apply the spin trajectory to single observations drawn from RealEstate10K, and compare our method using Ours-KF, Ours-1step, and Ours-Markov.
Our results are displayed in \reffig{fig:spin_fid_tsed}.
First, consider the per-frame FID (left inset):
    while all methods perform similarly close to the initial view,
    Ours-KF and Ours-1step exhibit parabolic behaviours, 
    where FID peaks and then decreases, even as error continues accumulating for Ours-Markov.
This is due to the cyclical nature of the path, 
    and the \textit{final} keyframe being conditioned on the \textit{initial} observation.
Views 
\begin{wrapfigure}{r}{0.49\linewidth}
    \centering
    \begin{subfigure}{0.49\linewidth}
        \pgfplotsset{width=1.3\linewidth,height=3.3cm,compat=1.18}
\begin{tikzpicture}
    \begin{axis}[
        title style={align=center, font=\scriptsize, yshift=-.5em},
        title={Spin\\Quality $\downarrow$},
        xlabel={Frame Index},
        ylabel={FID},
        xmin=0, xmax=9,
        ymin=50, ymax=80,
        xtick={0,1,2,3,4,5,6,7,8,9},
        ytick={20,40,60,80,100},
        legend pos=north west,
        legend style={nodes={scale=0.5, transform shape}},
        label style={font=\scriptsize},
        tick label style={font=\tiny},
        ymajorgrids=true,
        grid style=dashed,
        xlabel style={yshift=0.5ex},
        ylabel style={yshift=-1ex},
        mark size=0.8pt,
    ]
        \addplot[color=red,mark=square,] coordinates {
            (0, 52.25742636263334)(1, 53.62965857738092)(2, 55.32094447845844)(3, 58.48051781087551)(4, 60.42658385557212)(5, 61.61415188990193)(6, 63.018998962071066)(7, 64.59890976625573)(8, 65.36130400206906)(9, 66.8040692786029)
        };
        \addplot[color=black,mark=star,] coordinates {
            (0, 52.25742636263334)(1, 52.99779243173941)(2, 53.44947759229399)(3, 55.4030100252146)(4, 56.57368562508131)(5, 56.849282238239994)(6, 57.17955911077229)(7, 55.6057453235114)(8, 54.336875741070685)(9, 53.164951130004965)
        };
        \addplot[color=orange,mark=triangle,] coordinates {
            (0, 52.25742636263334)(1, 53.28469429111783)(2, 55.391211694370895)(3, 56.87982582768956)(4, 58.303136638167985)(5, 57.82915937618543)(6, 56.78107037396563)(7, 56.2874308464086)(8, 54.287079322033264)(9, 52.99028726935461)
        };
        \legend{Ours-Markov, Ours-1step, Ours-KF}
    \end{axis}
\end{tikzpicture}
    \end{subfigure}
    \hfill
    \begin{subfigure}{0.49\linewidth}
        \pgfplotsset{width=1.3\linewidth,height=3.3cm,compat=1.18}
\begin{tikzpicture}
    \begin{axis}[
        title style={align=center, font=\scriptsize, yshift=-.5em},
        title={Spin First-Last\\Consistency $\uparrow$},
        xlabel={$T_\text{error}$ (pixels)},
        ylabel={TSED},
        xmin=1, xmax=4,
        ymin=0, ymax=100,
        xtick={1,2,3,4},
        ytick={0,20,40,60,80,100},
        legend style={nodes={scale=0.5, transform shape},at={(0.03,0.5)},anchor=west},
        label style={font=\scriptsize},
        tick label style={font=\tiny},
        ymajorgrids=true,
        grid style=dashed,
        xlabel style={yshift=1ex},
        ylabel style={yshift=-2ex},
        mark size=0.5pt,
    ]
        \addplot[color=red,mark=o,] coordinates {
            (1.0,7.3999999999999995)(1.5,18.2)(2.0,28.000000000000004)(2.5,31.8)(3.0,34.8)(3.5,37.8)(4.0,40.400000000000006)
        };
        \addplot[color=black,mark=star,] coordinates {
            (1.0,89.0)(1.5,97.39999999999999)(2.0,97.6)(2.5,97.6)(3.0,98.0)(3.5,98.2)(4.0,98.2)
        };
        \addplot[color=orange,mark=triangle,] coordinates {
            (1.0,93.2)(1.5,98.0)(2.0,98.2)(2.5,98.8)(3.0,99.0)(3.5,99.0)(4.0,99.0)
        };
    \end{axis}
\end{tikzpicture}
    \end{subfigure}
    \caption{
        FID and TSED evaluation on RealEstate10K using the spin trajectory.
        For TSED, we examine the loop consistency of the first and last frames of the sequence.
    }
    \label{fig:spin_fid_tsed}
\end{wrapfigure}
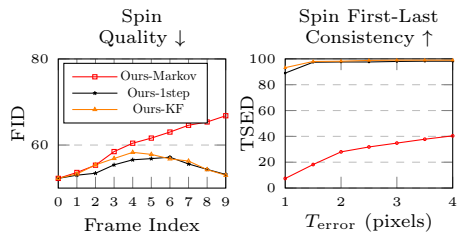towards the end of the spin are, therefore, able to incorporate information from the given view (with low depth, in the sense of \refsec{sec:samplinglargesets}), which results in better generation quality.
To analyze the loop consistency of synthesized views, we compute TSED on the first and last generated frames of the cyclical spin motion (see \reffig{fig:fridge} and \cite{Yu2023PhotoconsistentNVS}).
As expected, \reffig{fig:spin_fid_tsed} (right) shows that the
    set-based sampling methods are dramatically more effective at generating cyclical views-sets 
    (${>}90\%$ vs.\ ${<}40\%$ matching).
Qualitatively, the example shown in \reffig{fig:fridge} shows a region visible in the given image 
    that Ours-Markov fails to maintain, causing it to change in the final view.
Both Ours-KF and Ours-1step have high performance under cyclical settings, but Ours-KF shows that cycle consistency can be obtained without having to consider all views in one sampling step, which may be more practical.

\subsection{Grouped Sequential Generation}
Next, we consider NVS over sequences of groups of views, where views within a group cannot be naturally ordered; specifically, we consider trajectories of binocular stereo pairs.
We create these trajectories by treating the views of the original trajectories of RealEstate10K, as the ``right eye'' and adding a ``left'' one to create stereo pairs.
For a baseline, we apply standard autoregressive sampling with a \textit{zigzag} ordering, where the right and then the left views are sampled before repeating for the next stereo pair.
Our set-based approach avoids imposing such an ordering by performing \textit{grouped sampling} (a form of set-autoregression), where each set-generated stereo pair is conditioned on the previous pair.

Since we base our stereo trajectories on a real trajectory with ground-truth, we can perform reference-based evaluations on the \textit{right} view of each generated stereo pair.
Results in \reftab{tab:stereo_recon} are based on short-term and long-term views, as in \refsec{sec:seq_experiments}, which show that our grouped sampling outperforms the autoregressive method.
In addition,
we evaluate image quality over subsequent generations using FID.
The FID results in \reffig{fig:stereo_fid_tsed} are arranged using the same zigzag ordering used for the baseline.
Our grouped sampling significantly reduces the degradation of image quality compared to the simple autoregressive baseline.
\begin{figure}[t]
    \centering
    \begin{minipage}[b][][t]{.30\linewidth}
        \centering
        \captionof{table}{
            Reconstruction metrics on stereo pairs, where ground-truth reference frames are available (``right eye'' view).
        }
        \resizebox{\linewidth}{!}{
        \begin{tabular}{cc|c|cc}
            &&Method & LPIPS $\downarrow$ & PSNR $\uparrow$\\
             \hline
            \multirow{2}{*}{\rotatebox[origin=c]{90}{\resizebox{0.18\linewidth}{!}{Short-}}}  &
            \multirow{2}{*}{\rotatebox[origin=c]{90}{\resizebox{0.16\linewidth}{!}{term}}}  
                                                                                             &Ours-Markov       &        {0.550} &        {12.44}\\
                                                                                            & &Ours-Group        & \textbf{0.349} & \textbf{15.06}\\
             \hline
            \multirow{2}{*}{\rotatebox[origin=c]{90}{\resizebox{0.18\linewidth}{!}{Long-}}} & 
            \multirow{2}{*}{\rotatebox[origin=c]{90}{\resizebox{0.16\linewidth}{!}{term}}}  
                                                                                             &Ours-Markov       &        {0.722} &        {10.51} \\
                                                                                           &  &Ours-Group        & \textbf{0.649} & \textbf{11.35} \\
        \end{tabular}
        }
        \label{tab:stereo_recon}
    \end{minipage}
    \begin{minipage}[b][][t]{.69\linewidth}
        \begin{minipage}[b][][t]{.32\linewidth}
            \pgfplotsset{width=1.3\linewidth,height=3cm,compat=1.18}
            \begin{tikzpicture}
                \begin{axis}[
                    title style={align=center, font=\scriptsize,yshift=-.5em},
                    title={TSED Stereo\\Same $\uparrow$},
                    xlabel={$T_\text{error}$ (pixels)},
                    ylabel={TSED},
                    xmin=1, xmax=4,
                    ymin=0, ymax=100,
                    xtick={},
                    ytick={0,50,100},
                    legend pos=north west,
                    legend style={nodes={scale=0.5, transform shape}},
                    label style={font=\scriptsize},
                    tick label style={font=\tiny},
                    ymajorgrids=true,
                    grid style=dashed,
                    xlabel style={yshift=1ex},
                    ylabel style={yshift=-2ex},
                ]
                    \addplot[color=orange,mark=square,] coordinates {
                        (1.0,76.52794653705953)(1.5,87.92527339003644)(2.0,91.88639125151884)(2.5,93.56622114216283)(3.0,94.53827460510327)(3.5,95.10783718104496)(4.0,95.48298906439854)
                    };
                    \addplot[color=red,mark=triangle,] coordinates {
                        (1.0,6.471749696233293)(1.5,14.410692588092346)(2.0,21.754252733900366)(2.5,28.12120291616039)(3.0,33.45382746051033)(3.5,37.93894289185906)(4.0,41.97904009720535)
                    };
                \end{axis}
            \end{tikzpicture}
        \end{minipage}
        \begin{minipage}[b][][t]{.32\linewidth}
            \pgfplotsset{width=1.3\linewidth,height=3cm,compat=1.18}
            \begin{tikzpicture}
                \begin{axis}[
                    title style={align=center, font=\scriptsize,yshift=-.5em},
                    title={TSED Stereo\\Diag $\uparrow$},
                    xlabel={$T_\text{error}$ (pixels)},
                    ylabel={TSED},
                    xmin=1, xmax=4,
                    ymin=0, ymax=100,
                    xtick={1,2,3,4},
                    ytick={0,50,100},
                    legend pos=south west,
                    legend style={nodes={scale=0.5, transform shape}},
                    label style={font=\scriptsize},
                    tick label style={font=\tiny},
                    ymajorgrids=true,
                    grid style=dashed,
                    xlabel style={yshift=1ex},
                    ylabel style={yshift=-2ex},
                ]
                    \addplot[color=orange,mark=square,] coordinates {
                        (1.0,67.38456865127581)(1.5,79.43955042527338)(2.0,83.8350546780073)(2.5,85.80650060753342)(3.0,86.78918590522478)(3.5,87.4179829890644)(4.0,87.80224787363305)
                    };
                    \addplot[color=red,mark=triangle,] coordinates {
                        (1.0,48.576852976913734)(1.5,63.066524908869994)(2.0,69.4198055893074)(2.5,72.92982989064397)(3.0,75.11239368165249)(3.5,76.47630619684081)(4.0,77.43620899149454)
                    };
                    \legend{Ours-Group, Ours-Markov}
                \end{axis}
            \end{tikzpicture}
        \end{minipage}
        \begin{minipage}[b][][t]{.32\linewidth}
            \pgfplotsset{width=1.3\linewidth,height=3cm,compat=1.18}
            \begin{tikzpicture}
                \begin{axis}[
                    title style={align=center, font=\scriptsize,yshift=-.5em},
                    title={FID\\Stereo $\downarrow$},
                    xlabel={Frame Index},
                    ylabel={FID},
                    xmin=0, xmax=42,
                    ymin=20, ymax=80,
                    xtick={0,20,40},
                    ytick={0,10,20,40,60,80},
                    legend pos=north west,
                    legend style={nodes={scale=0.5, transform shape}},
                    label style={font=\scriptsize},
                    tick label style={font=\tiny},
                    ymajorgrids=true,
                    grid style=dashed,
                    xlabel style={yshift=1ex},
                    ylabel style={yshift=-1ex},
                    mark size=\marksize,
                ]
                    \addplot[color=orange,mark=square,] coordinates {
                        (0, 23.19077917704658)(1, 25.279293407791727)(2, 23.421675110820956)(3, 25.109601302389592)(4, 24.538187146160737)(5, 26.173056887847622)(6, 25.524146170884592)(7, 27.339684165821495)(8, 27.063788721742583)(9, 28.84718663555833)(10, 28.496280948306207)(11, 30.287261602155525)(12, 30.361056367121535)(13, 32.15141572523834)(14, 32.31662748462185)(15, 33.81380569670921)(16, 34.22117034138455)(17, 35.92050939345859)(18, 36.30448345547222)(19, 38.230582132981226)(20, 38.52994144789187)(21, 40.354303086066125)(22, 40.59991386240148)(23, 42.84957968037094)(24, 42.855388837187434)(25, 45.46151879978339)(26, 45.88448086334529)(27, 47.538956947662996)(28, 47.84708848711506)(29, 50.12483713740198)(30, 50.31975940384564)(31, 52.53138259504368)(32, 53.2491170925166)(33, 54.51111242033437)(34, 55.027325352170635)(35, 57.24347129014848)(36, 56.83367747012801)(37, 59.481866252729105)(38, 59.187971274033885)(39, 62.05954936891766)(40, 61.916372797533256)(41, 63.25317929168307)
                    };
                    \addplot[color=red,mark=triangle,] coordinates {
                        (0, 23.19175319193772)(1, 25.54450646521991)(2, 27.040811427199515)(3, 28.8324394211632)(4, 29.65364192287302)(5, 31.56338740463974)(6, 32.64879686165693)(7, 35.11638337352747)(8, 36.12088487746519)(9, 38.761613512999816)(10, 39.13309973184937)(11, 41.37676474329521)(12, 41.921760038586115)(13, 44.291151293804546)(14, 44.69264711234979)(15, 47.359407469091934)(16, 48.897078518703836)(17, 51.10910625647176)(18, 51.993432827131585)(19, 54.29130377790523)(20, 55.297169742763174)(21, 56.42362183716608)(22, 56.930275129366294)(23, 58.50835291004364)(24, 60.19912434487361)(25, 61.906792704746636)(26, 62.384888470037026)(27, 64.27366825910406)(28, 64.44454990159903)(29, 66.27768657931159)(30, 67.25038870687649)(31, 69.58311478938606)(32, 69.31977591728827)(33, 70.93184881822924)(34, 71.70956057005094)(35, 72.36574259788983)(36, 72.25836471733066)(37, 73.22821993367)(38, 73.69771034000274)(39, 74.69718727544677)(40, 75.70751580462576)(41, 76.98223004288116)
                    };
                \end{axis}
            \end{tikzpicture}
        \end{minipage}
        \captionof{figure}{
            TSED plots measuring inter-frame consistency between same-eye generations (left) and different-eye generations (middle), across adjacent stereo pairs in time.
            FID across time (right), ordered in the ``zigzag'' manner.
        }
        \label{fig:stereo_fid_tsed}
    \end{minipage}
\end{figure}

\begin{figure}[t]
    \centering
    \includegraphics[width=\linewidth]{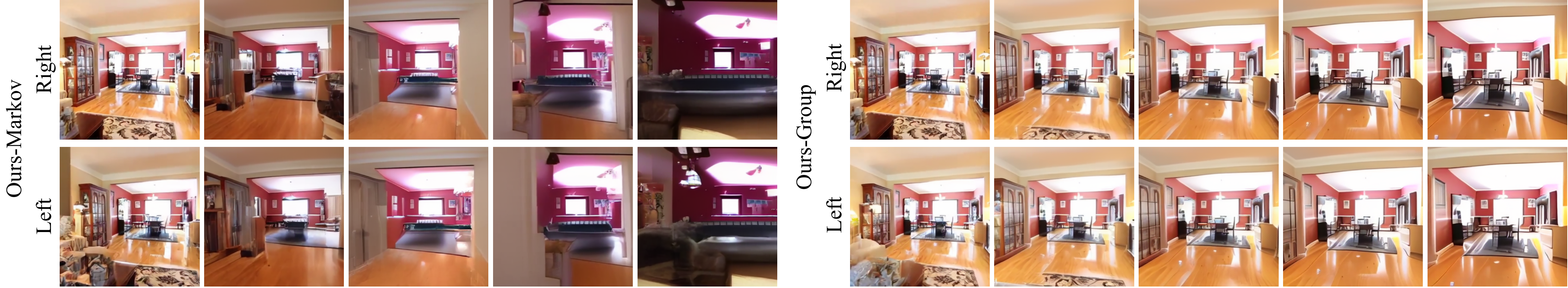}
    \caption{
        A qualitative example of a generated stereo pair using standard autoregressive and grouped sampling.
        Notice that the scene content diverges quickly when using standard autoregressive generation.
        Views progress from the left to the right columns.
    }
    \label{fig:stereo_qual}
\end{figure}

Finally, we use TSED to evaluate consistency between generations. 
However, the lack of a natural order overall views requires careful selection of the pairs used for evaluation.
We therefore consider two types of view pairs for TSED evaluations: 
    same-sided and cross-sided.
For same-sided evaluations, pairs of views from the same side of the stereo pairs (i.e., left or right), and adjacent in the sequence of pairs are considered.
For cross-sided evaluations, pairs on the opposite sides of the stereo pairs, and adjacent in the sequence of pairs, are considered.
The TSED results in \reffig{fig:stereo_fid_tsed} show that grouped (i.e., set-wise) generation is significantly more consistent, especially for the same-sided evaluations.
This consistency can be observed qualitatively in \reffig{fig:stereo_qual}.
Overall, our set-based grouped method, which is more strongly constrained by multi-view conditioning and does not impose an arbitrary ordering on the binocular pairs, results in better generations in terms of image quality and cross-frame consistency.

\section{Conclusion}
In this paper, we presented a set-based approach for generative novel view synthesis (GNVS). 
Not only are many camera view-sets of interest inherently difficult to order, but set-based generation can help alleviate the accumulation of errors in autoregressive generation.
To this end, we devised a flexible set-to-set generative model which both conditions on and generates \textit{sets} of images, in a permutation-invariant manner.
We evaluated the model using standard NVS metrics and datasets, obtaining improved image quality.
We also demonstrated even larger improvements when generating particularly challenging view-sets, including cyclic paths (which induce loop inconsistencies in prior methods) and binocular trajectories (which are not naturally ordered).
We believe that this approach can help mitigate common problems with image-based GNVS, and form the foundation for future GNVS techniques generally.

%
%
\clearpage
\section*{Acknowledgements}
This work was completed with support from the Vector Institute, and was funded in part by the Canada First Research Excellence Fund (CFREF) for the Vision: Science to Applications (VISTA) program (M.A.B., K.G.D., T.T.A.A.), the NSERC Discovery Grant program (M.A.B., K.G.D.), and the NSERC Canada Graduate Scholarship Doctoral program (J.J.Y.).

\bibliographystyle{splncs04}
\bibliography{refs_short,refs}

\begin{thebibliography}{10}
\providecommand{\url}[1]{\texttt{#1}}
\providecommand{\urlprefix}{URL }
\providecommand{\doi}[1]{https://doi.org/#1}

\bibitem{anciukevicius2022renderdiffusion}
Anciukevicius, T., Xu, Z., Fisher, M., Henderson, P., Bilen, H., Mitra, N.J., Guerrero, P.: {RenderDiffusion}: Image diffusion for {3D} reconstruction, inpainting and generation. {Proceedings of the {IEEE} Conference on Computer Vision and Pattern Recognition ({CVPR})}  (2023)

\bibitem{aoki2019pointnetlk}
Aoki, Y., Goforth, H., Srivatsan, R.A., Lucey, S.: {PointNetLK}: Robust \& efficient point cloud registration using {PointNet}. In: {Proceedings of the {IEEE} Conference on Computer Vision and Pattern Recognition ({CVPR})} (2019)

\bibitem{avidan1997novel}
Avidan, S., Shashua, A.: Novel view synthesis in tensor space. In: {Proceedings of the {IEEE} Conference on Computer Vision and Pattern Recognition ({CVPR})} (1997)

\bibitem{barron2023zip}
Barron, J.T., Mildenhall, B., Verbin, D., Srinivasan, P.P., Hedman, P.: {Zip-NeRF}: Anti-aliased grid-based neural radiance fields. {Proceedings of the International Conference on Computer Vision ({ICCV})}  (2023)

\bibitem{bautista2022gaudi}
Bautista, M.{\'A}., Guo, P., Abnar, S., Talbott, W., Toshev, A.T., Chen, Z., Dinh, L., Zhai, S., Goh, H., Ulbricht, D., Dehghan, A., Susskind, J.M.: {GAUDI}: A neural architect for immersive 3{D} scene generation. In: {Neural Information Processing Systems ({NeurIPS})} (2022)

\bibitem{blattmann2023align}
Blattmann, A., Rombach, R., Ling, H., Dockhorn, T., Kim, S.W., Fidler, S., Kreis, K.: Align your latents: High-resolution video synthesis with latent diffusion models. In: {Proceedings of the {IEEE} Conference on Computer Vision and Pattern Recognition ({CVPR})} (2023)

\bibitem{carion2020end}
Carion, N., Massa, F., Synnaeve, G., Usunier, N., Kirillov, A., Zagoruyko, S.: End-to-end object detection with transformers. In: {Proceedings of the European Conference on Computer Vision ({ECCV})} (2020)

\bibitem{chan2023genvs}
Chan, E.R., Nagano, K., Chan, M.A., Bergman, A.W., Park, J.J., Levy, A., Aittala, M., Mello, S.D., Karras, T., Wetzstein, G.: {GeNVS}: Generative novel view synthesis with {3D}-aware diffusion models. In: {Proceedings of the International Conference on Computer Vision ({ICCV})} (2023)

\bibitem{chan2007image}
Chan, S., Shum, H.Y., Ng, K.T.: Image-based rendering and synthesis. IEEE Signal Processing Magazine  (2007)

\bibitem{chang2017matterport3d}
Chang, A., Dai, A., Funkhouser, T., Halber, M., Niessner, M., Savva, M., Song, S., Zeng, A., Zhang, Y.: Matterport3{D}: Learning from {RGB-D} data in indoor environments. {Proceedings of the International Conference on 3D Vision ({3DV})}  (2017)

\bibitem{chen1993view}
Chen, S.E., Williams, L.: View interpolation for image synthesis. In: {Proceedings of {SIGGRAPH}} (1993)

\bibitem{croitoru2023diffusion}
Croitoru, F.A., Hondru, V., Ionescu, R.T., Shah, M.: Diffusion models in vision: A survey. {{IEEE} Transactions on Pattern Analysis and Machine Intelligence ({PAMI})}  (2023)

\bibitem{deitke2023objaverse}
Deitke, M., Liu, R., Wallingford, M., Ngo, H., Michel, O., Kusupati, A., Fan, A., Laforte, C., Voleti, V., Gadre, S.Y., VanderBilt, E., Kembhavi, A., Vondrick, C., Gkioxari, G., Ehsani, K., Schmidt, L., Farhadi, A.: {Objaverse}-{XL}: A universe of {10M+} {3D} objects. {Neural Information Processing Systems ({NeurIPS})}  (2023)

\bibitem{esser2021taming}
Esser, P., Rombach, R., Ommer, B.: Taming transformers for high-resolution image synthesis. In: {Proceedings of the {IEEE} Conference on Computer Vision and Pattern Recognition ({CVPR})} (2021)

\bibitem{flynn2019deepview}
Flynn, J., Broxton, M., Debevec, P., DuVall, M., Fyffe, G., Overbeck, R., Snavely, N., Tucker, R.: {DeepView}: View synthesis with learned gradient descent. In: {Proceedings of the {IEEE} Conference on Computer Vision and Pattern Recognition ({CVPR})} (2019)

\bibitem{gao2022nerf}
Gao, K., Gao, Y., He, H., Lu, D., Xu, L., Li, J.: {NeRF}: Neural radiance field in {3D} vision, a comprehensive review. arXiv preprint arXiv:2210.00379  (2022)

\bibitem{geyer2023tokenflow}
Geyer, M., Bar-Tal, O., Bagon, S., Dekel, T.: {TokenFlow}: Consistent diffusion features for consistent video editing. {Proceedings of the International Conference on Learning Representations ({ICLR})}  (2024)

\bibitem{he2022voxel}
He, C., Li, R., Li, S., Zhang, L.: Voxel set transformer: A set-to-set approach to {3D} object detection from point clouds. In: {Proceedings of the {IEEE} Conference on Computer Vision and Pattern Recognition ({CVPR})} (2022)

\bibitem{heusel2017gans}
Heusel, M., Ramsauer, H., Unterthiner, T., Nessler, B., Hochreiter, S.: {GANs} trained by a two time-scale update rule converge to a local {Nash} equilibrium. {Neural Information Processing Systems ({NeurIPS})}  (2017)

\bibitem{ho2020denoising}
Ho, J., Jain, A., Abbeel, P.: Denoising diffusion probabilistic models. {Neural Information Processing Systems ({NeurIPS})}  (2020)

\bibitem{ho2022video}
Ho, J., Salimans, T., Gritsenko, A., Chan, W., Norouzi, M., Fleet, D.J.: Video diffusion models. {Proceedings of the International Conference on Learning Representations ({ICLR})}  (2022)

\bibitem{kim2021setvae}
Kim, J., Yoo, J., Lee, J., Hong, S.: {SetVAE}: Learning hierarchical composition for generative modeling of set-structured data. In: {Proceedings of the {IEEE} Conference on Computer Vision and Pattern Recognition ({CVPR})} (2021)

\bibitem{kim2023neuralfield}
Kim, S.W., Brown, B., Yin, K., Kreis, K., Schwarz, K., Li, D., Rombach, R., Torralba, A., Fidler, S.: {NeuralField-LDM}: Scene generation with hierarchical latent diffusion models. In: {Proceedings of the {IEEE} Conference on Computer Vision and Pattern Recognition ({CVPR})} (2023)

\bibitem{laveau19943}
Laveau, S., Faugeras, O.D.: {3-D} scene representation as a collection of images. In: {Proceedings of the International Conference on Pattern Recognition ({ICPR})} (1994)

\bibitem{lee2019set}
Lee, J., Lee, Y., Kim, J., Kosiorek, A., Choi, S., Teh, Y.W.: Set transformer: A framework for attention-based permutation-invariant neural networks. In: {Proceedings of the International Conference on Machine Learning ({ICML})} (2019)

\bibitem{liu2021infinite}
Liu, A., Tucker, R., Jampani, V., Makadia, A., Snavely, N., Kanazawa, A.: Infinite nature: Perpetual view generation of natural scenes from a single image. In: {Proceedings of the International Conference on Computer Vision ({ICCV})} (2021)

\bibitem{liu2023zero1to3}
Liu, R., Wu, R., Hoorick, B.V., Tokmakov, P., Zakharov, S., Vondrick, C.: {Zero-1-to-3}: Zero-shot one image to {3D} object. In: {Proceedings of the International Conference on Computer Vision ({ICCV})} (2023)

\bibitem{lowe1999object}
Lowe, D.G.: Object recognition from local scale-invariant features. In: {Proceedings of the International Conference on Computer Vision ({ICCV})} (1999)

\bibitem{luo2021diffusion}
Luo, S., Hu, W.: Diffusion probabilistic models for {3D} point cloud generation. In: {Proceedings of the {IEEE} Conference on Computer Vision and Pattern Recognition ({CVPR})} (2021)

\bibitem{mildenhall2021nerf}
Mildenhall, B., Srinivasan, P.P., Tancik, M., Barron, J.T., Ramamoorthi, R., Ng, R.: {NeRF}: Representing scenes as neural radiance fields for view synthesis. In: {Proceedings of the European Conference on Computer Vision ({ECCV})} (2020)

\bibitem{NEURIPS2021_701d8045}
Nie, W., Vahdat, A., Anandkumar, A.: Controllable and compositional generation with latent-space energy-based models. In: {Neural Information Processing Systems ({NeurIPS})} (2021)

\bibitem{po2023state}
Po, R., Yifan, W., Golyanik, V., Aberman, K., Barron, J.T., Bermano, A.H., Chan, E.R., Dekel, T., Holynski, A., Kanazawa, A., Liu, C.K., Liu, L., Mildenhall, B., Nießner, M., Ommer, B., Theobalt, C., Wonka, P., Wetzstein, G.: State of the art on diffusion models for visual computing. arXiv preprint arXiv:2310.07204  (2023)

\bibitem{qi2017pointnet}
Qi, C.R., Su, H., Mo, K., Guibas, L.J.: {PointNet}: Deep learning on point sets for {3D} classification and segmentation. In: {Proceedings of the {IEEE} Conference on Computer Vision and Pattern Recognition ({CVPR})} (2017)

\bibitem{qi2017pointnet++}
Qi, C.R., Yi, L., Su, H., Guibas, L.J.: {PointNet++}: Deep hierarchical feature learning on point sets in a metric space. {Neural Information Processing Systems ({NeurIPS})}  (2017)

\bibitem{ravanbakhsh2016deep}
Ravanbakhsh, S., Schneider, J., Poczos, B.: Deep learning with sets and point clouds. {Proceedings of the International Conference on Learning Representations ({ICLR})}  (2016)

\bibitem{ren2022look}
Ren, X., Wang, X.: Look outside the room: Synthesizing a consistent long-term {3D} scene video from a single image. In: {Proceedings of the {IEEE} Conference on Computer Vision and Pattern Recognition ({CVPR})} (2022)

\bibitem{rombach2022high}
Rombach, R., Blattmann, A., Lorenz, D., Esser, P., Ommer, B.: High-resolution image synthesis with latent diffusion models. In: {Proceedings of the {IEEE} Conference on Computer Vision and Pattern Recognition ({CVPR})} (2022)

\bibitem{rombach2021geometry}
Rombach, R., Esser, P., Ommer, B.: Geometry-free view synthesis: Transformers and no {3D} priors. In: {Proceedings of the International Conference on Computer Vision ({ICCV})} (2021)

\bibitem{ronneberger2015u}
Ronneberger, O., Fischer, P., Brox, T.: U-{N}et: Convolutional networks for biomedical image segmentation. In: {Proceedings of the International Conference on Medical Image Computing and Computer Assisted Intervention ({MICCAI})} (2015)

\bibitem{sajjadi2022scene}
Sajjadi, M.S.M., Meyer, H., Pot, E., Bergmann, U., Greff, K., Radwan, N., Vora, S., Lucic, M., Duckworth, D., Dosovitskiy, A., Uszkoreit, J., Funkhouser, T., Tagliasacchi, A.: Scene representation transformer: Geometry-free novel view synthesis through set-latent scene representations. In: {Proceedings of the {IEEE} Conference on Computer Vision and Pattern Recognition ({CVPR})} (2022)

\bibitem{sarlin2020superglue}
Sarlin, P.E., DeTone, D., Malisiewicz, T., Rabinovich, A.: {SuperGlue}: Learning feature matching with graph neural networks. In: {Proceedings of the {IEEE} Conference on Computer Vision and Pattern Recognition ({CVPR})} (2020)

\bibitem{savva2019habitat}
Savva, M., Kadian, A., Maksymets, O., Zhao, Y., Wijmans, E., Jain, B., Straub, J., Liu, J., Koltun, V., Malik, J., Parikh, D., Batra, D.: Habitat: A platform for embodied {AI} research. In: {Proceedings of the International Conference on Computer Vision ({ICCV})} (2019)

\bibitem{scharstein1996stereo}
Scharstein, D.: Stereo vision for view synthesis. In: {Proceedings of the {IEEE} Conference on Computer Vision and Pattern Recognition ({CVPR})} (1996)

\bibitem{segol2019universal}
Segol, N., Lipman, Y.: On universal equivariant set networks. {Proceedings of the International Conference on Learning Representations ({ICLR})}  (2020)

\bibitem{seitz1995physically}
Seitz, S.M., Dyer, C.R.: Physically-valid view synthesis by image interpolation. In: ICCV Workshop on Representation of Visual Scenes (1995)

\bibitem{shum2000review}
Shum, H., Kang, S.B.: Review of image-based rendering techniques. In: Visual Communications and Image Processing. SPIE (2000)

\bibitem{sohl2015deep}
Sohl-Dickstein, J., Weiss, E., Maheswaranathan, N., Ganguli, S.: Deep unsupervised learning using nonequilibrium thermodynamics. In: {Proceedings of the International Conference on Machine Learning ({ICML})} (2015)

\bibitem{song2020denoising}
Song, J., Meng, C., Ermon, S.: Denoising diffusion implicit models. {Proceedings of the International Conference on Learning Representations ({ICLR})}  (2021)

\bibitem{szymanowicz23viewset_diffusion}
Szymanowicz, S., Rupprecht, C., Vedaldi, A.: Viewset diffusion: (0-)image-conditioned {3D} generative models from {2D} data. In: {Proceedings of the International Conference on Computer Vision ({ICCV})} (2023)

\bibitem{tancik2020fourier}
Tancik, M., Srinivasan, P., Mildenhall, B., Fridovich-Keil, S., Raghavan, N., Singhal, U., Ramamoorthi, R., Barron, J., Ng, R.: Fourier features let networks learn high frequency functions in low dimensional domains. {Neural Information Processing Systems ({NeurIPS})}  (2020)

\bibitem{tewari2023diffusion}
Tewari, A., Yin, T., Cazenavette, G., Rezchikov, S., Tenenbaum, J.B., Durand, F., Freeman, W.T., Sitzmann, V.: Diffusion with forward models: Solving stochastic inverse problems without direct supervision. {Neural Information Processing Systems ({NeurIPS})}  (2023)

\bibitem{poseguideddiffusion}
Tseng, H.Y., Li, Q., Kim, C., Alsisan, S., Huang, J.B., Kopf, J.: Consistent view synthesis with pose-guided diffusion models. In: {Proceedings of the {IEEE} Conference on Computer Vision and Pattern Recognition ({CVPR})} (2023)

\bibitem{vaswani2017attention}
Vaswani, A., Shazeer, N., Parmar, N., Uszkoreit, J., Jones, L., Gomez, A.N., Kaiser, {\L}., Polosukhin, I.: Attention is all you need. {Neural Information Processing Systems ({NeurIPS})}  (2017)

\bibitem{wang2021ibrnet}
Wang, Q., Wang, Z., Genova, K., Srinivasan, P.P., Zhou, H., Barron, J.T., Martin-Brualla, R., Snavely, N., Funkhouser, T.: {IBRNet}: Learning multi-view image-based rendering. In: {Proceedings of the {IEEE} Conference on Computer Vision and Pattern Recognition ({CVPR})} (2021)

\bibitem{watson2022novel}
Watson, D., Chan, W., Martin-Brualla, R., Ho, J., Tagliasacchi, A., Norouzi, M.: Novel view synthesis with diffusion models. {Proceedings of the International Conference on Learning Representations ({ICLR})}  (2023)

\bibitem{xie2022neural}
Xie, Y., Takikawa, T., Saito, S., Litany, O., Yan, S., Khan, N., Tombari, F., Tompkin, J., Sitzmann, V., Sridhar, S.: Neural fields in visual computing and beyond. In: Computer Graphics Forum (2022)

\bibitem{yogamani2019woodscape}
Yogamani, S., Hughes, C., Horgan, J., Sistu, G., Varley, P., O'Dea, D., Uric{\'a}r, M., Milz, S., Simon, M., Amende, K., Witt, C., Rashed, H., Chennupati, S., Nayak, S., Mansoor, S., Perroton, X., Perez, P.: {WoodScape}: A multi-task, multi-camera fisheye dataset for autonomous driving. In: {Proceedings of the International Conference on Computer Vision ({ICCV})} (2019)

\bibitem{yu2021pixelnerf}
Yu, A., Ye, V., Tancik, M., Kanazawa, A.: {pixelNeRF}: Neural radiance fields from one or few images. In: {Proceedings of the {IEEE} Conference on Computer Vision and Pattern Recognition ({CVPR})} (2021)

\bibitem{Yu2023PhotoconsistentNVS}
Yu, J.J., Forghani, F., Derpanis, K.G., Brubaker, M.A.: Long-term photometric consistent novel view synthesis with diffusion models. In: {Proceedings of the International Conference on Computer Vision ({ICCV})} (2023)

\bibitem{zaheer2017deep}
Zaheer, M., Kottur, S., Ravanbakhsh, S., Poczos, B., Salakhutdinov, R.R., Smola, A.J.: Deep sets. {Neural Information Processing Systems ({NeurIPS})}  (2017)

\bibitem{zhang2018unreasonable}
Zhang, R., Isola, P., Efros, A.A., Shechtman, E., Wang, O.: The unreasonable effectiveness of deep features as a perceptual metric. In: {Proceedings of the {IEEE} Conference on Computer Vision and Pattern Recognition ({CVPR})} (2018)

\bibitem{zhang2019deep}
Zhang, Y., Hare, J., Prugel-Bennett, A.: Deep set prediction networks. {Neural Information Processing Systems ({NeurIPS})}  (2019)

\bibitem{zhang2019fspool}
Zhang, Y., Hare, J., Pr{\"u}gel-Bennett, A.: {FSPool}: Learning set representations with featurewise sort pooling. In: {Proceedings of the International Conference on Learning Representations ({ICLR})} (2019)

\bibitem{zhang2021multiset}
Zhang, Y., Zhang, D.W., Lacoste-Julien, S., Burghouts, G.J., Snoek, C.G.: Multiset-equivariant set prediction with approximate implicit differentiation. {Proceedings of the International Conference on Learning Representations ({ICLR})}  (2022)

\bibitem{zhou2018stereo}
Zhou, T., Tucker, R., Flynn, J., Fyffe, G., Snavely, N.: Stereo magnification. ACM Transactions on Graphics  (2018)

\end{thebibliography}
\clearpage

\appendix

\section{Camera Ray Encoding}
\label{sec:rays}
To provide our model with access to the camera geometry, we adopt the ray representation used in previous works \cite{Yu2023PhotoconsistentNVS,sajjadi2022scene}.
Given the intrinsic matrix, $\mathbf{K}$, and the extrinsic matrix, $\left[\mathbf{R}|\mathbf{t}\right]$, of a camera, the projection matrix is defined as $\mathbf{P} = \mathbf{K}[\mathbf{R}|\mathbf{t}]$.
The ray $\ray_{u,v} = (\boldsymbol{\tau},\dir_{u,v})$ at pixel coordinates $(u,v)$ is composed of the camera center $\boldsymbol{\tau} = -\mathbf{R}^{-1}\mathbf{t}$, and normalized direction $\dir_{u,v}$.
The unnormalized ray direction is given by:
\begin{align}
    \Bar{\dir}_{u,v} = \mathbf{R}^{-1} \mathbf{K}^{-1}
    \begin{bmatrix}
        u & v & 1
    \end{bmatrix}^\top.
\end{align}
The rays, $\ray$, are frequency encoded \cite{vaswani2017attention} into their final representation, $\mathcal{R}$, which is then used to condition the model:
\begin{align}
    \mathcal{R} = \left[ \sin(f_1 \pi \ray), \cos(f_1 \pi \ray), \dotsc, \sin(f_K \pi \ray), \cos(f_K \pi \ray) \right],
\end{align}
where $K$ frequencies are used, which increment in frequency by powers of two.

\section{Generalization to Number of Views}
\label{sec:general}
Computational constraints limit the number of views that can be used during training.
In this section, we provide additional evaluations that investigate the impact of sampling using more views than those seen by our model during training.

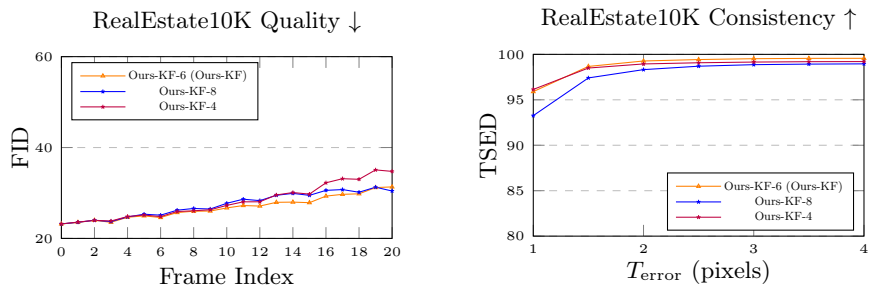
\begin{figure}
    \centering
    \begin{subfigure}{0.49\linewidth}
        \pgfplotsset{width=\linewidth,height=4cm,compat=1.18}
        \begin{tikzpicture}
            \begin{axis}[
                title={RealEstate10K Quality $\downarrow$},
                xlabel={Frame Index},
                ylabel={FID},
                xmin=0, xmax=20,
                ymin=20, ymax=60,
                xtick={0,2,4,6,8,10,12,14,16,18,20},
                ytick={0,20,40,60,80,100},
                legend pos=north west,
                legend style={nodes={scale=0.5, transform shape}},
                label style={font=\small},
                tick label style={font=\tiny},
                ymajorgrids=true,
                grid style=dashed,
                xlabel style={yshift=0.5ex},
                ylabel style={yshift=-1ex},
                mark size=0.8pt,
            ]
                \addplot[color=orange,mark=triangle,] coordinates {
                    (0, 23.191309803640422)(1, 23.51614990681969)(2, 23.990226908808324)(3, 23.543027198182813)(4, 24.652106973566788)(5, 24.960601465382865)(6, 24.595232241333804)(7, 25.690130414383248)(8, 25.98251904596549)(9, 26.022328561151085)(10, 26.707767623294785)(11, 27.23416156086023)(12, 27.11967559177299)(13, 27.944219965430307)(14, 27.988278344121113)(15, 27.86203308760895)(16, 29.31409494242274)(17, 29.695871862108334)(18, 29.816069114379275)(19, 31.18597657793879)(20, 31.336071246278948)
                };
                \addplot[color=blue,mark=star,] coordinates {
                    (0, 23.191309803640422)(1, 23.552564069748314)(2, 23.989303969781417)(3, 23.71037798259374)(4, 24.722430325441906)(5, 25.323697766260125)(6, 25.115631327264282)(7, 26.211905639186114)(8, 26.59025552114906)(9, 26.471883474101105)(10, 27.724169249407396)(11, 28.632497287787885)(12, 28.312109086810494)(13, 29.49043647470762)(14, 29.906990151026207)(15, 29.478072134068043)(16, 30.571422157628547)(17, 30.748080813827585)(18, 30.160154392913853)(19, 31.272415870553573)(20, 30.42773587888098)
                };
                \addplot[color=purple,mark=star,] coordinates {
                    (0, 23.191309803640422)(1, 23.59305906338284)(2, 23.98266726844065)(3, 23.829641054027945)(4, 24.817614478749306)(5, 25.141235332669794)(6, 24.8009816604619)(7, 25.919986199529262)(8, 26.076916753990872)(9, 26.305904979383)(10, 27.26372470480692)(11, 28.05012451817987)(12, 28.061382544852336)(13, 29.568368915261146)(14, 30.091467626806434)(15, 29.74928815682574)(16, 32.23994608634479)(17, 33.13946417139198)(18, 33.00851979138071)(19, 35.07666539387037)(20, 34.74935459383346)
                };
                \legend{Ours-KF-6 (Ours-KF), Ours-KF-8, Ours-KF-4}
            \end{axis}
        \end{tikzpicture}
    \end{subfigure}
    \hfill
    \begin{subfigure}{0.49\linewidth}
        \pgfplotsset{width=\linewidth,height=4cm,compat=1.18}
        \begin{tikzpicture}
            \begin{axis}[
                title={RealEstate10K Consistency $\uparrow$},
                xlabel={$T_\text{error}$ (pixels)},
                ylabel={TSED},
                xmin=1, xmax=4,
                ymin=80, ymax=100,
                xtick={1,2,3,4},
                ytick={80, 85,90,95,100},
                legend pos=south east,
                legend style={nodes={scale=0.5, transform shape}},
                label style={font=\small},
                tick label style={font=\tiny},
                ymajorgrids=true,
                grid style=dashed,
                xlabel style={yshift=1ex},
                ylabel style={yshift=-1.5ex},
                mark size=0.8pt,
            ]
                \addplot[color=orange,mark=triangle,] coordinates {
                    (1.0,95.90218712029161)(1.5,98.67253948967193)(2.0,99.27095990279466)(2.5,99.42891859052247)(3.0,99.52308626974484)(3.5,99.5595382746051)(4.0,99.56561360874849)
                };
                \addplot[color=blue,mark=star,] coordinates {
                    (1.0,93.25941676792225)(1.5,97.40886998784933)(2.0,98.32017010935601)(2.5,98.69684082624545)(3.0,98.86695018226001)(3.5,98.9277035236938)(4.0,98.955042527339)
                };
                \addplot[color=purple,mark=star,] coordinates {
                    (1.0,96.14216281895504)(1.5,98.49635479951398)(2.0,98.94592952612393)(2.5,99.07654921020657)(3.0,99.15249088699879)(3.5,99.19501822600243)(4.0,99.20716889428918)
                };
                \legend{Ours-KF-6 (Ours-KF), Ours-KF-8, Ours-KF-4}
            \end{axis}
        \end{tikzpicture}
    \end{subfigure}
    \caption{
        Quality and consistency of our generations on RealEstate10k while varying the number total number of conditioning and generating views.
        $\text{Ours-KF-}i$ denotes generation using keyframes where the maximum number of total views used for simultaneous generation is at most $i$.
        In all cases, the maximum number of conditioning views is $\frac{i}{2}$.
        Ours-KF-6 is the model presented in the main paper as Ours-KF.
        Note that performance is best using Ours-KF-6, which does not use a significantly larger number of views as seen during training, which would lower generated image quality as seen with Ours-KF-8, while having a lower \textit{sampling depth} than Ours-KF-4, which reduces error accumulation from autoregressive generation.
    }
    \label{fig:ood_n_views}
\end{figure}

First, we investigate the impact of the maximum \textit{total} number of views used for our keyframing sampling approach.
The total number of views is the sum of the number of conditioning and generated views.
The keyframing results presented in the main paper use a maximum total of six (combined). In this section, we investigate higher and lower maximum totals of eight and four.
In all cases, the maximum number of conditioning views is half that of the total number of views.
The results shown in \reffig{fig:ood_n_views} show that increasing the maximum total views reduces both generated image quality (see, e.g., the FID of frames 12-16) and TSED consistency (especially for $T_\text{error} \leq 2$), while lowering the limit also reduces quality (see FID on frame 12 and higher), while slightly lowering consistency.

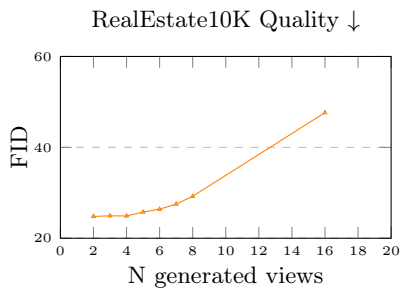
\begin{figure}
    \centering
    \begin{subfigure}{0.49\linewidth}
        \pgfplotsset{width=\linewidth,height=4cm,compat=1.18}
        \begin{tikzpicture}
            \begin{axis}[
                title={RealEstate10K Quality $\downarrow$},
                xlabel={N generated views},
                ylabel={FID},
                xmin=0, xmax=20,
                ymin=20, ymax=60,
                xtick={0,2,4,6,8,10,12,14,16,18,20},
                ytick={0,20,40,60,80,100},
                legend pos=north west,
                legend style={nodes={scale=0.5, transform shape}},
                label style={font=\small},
                tick label style={font=\tiny},
                ymajorgrids=true,
                grid style=dashed,
                xlabel style={yshift=0.5ex},
                ylabel style={yshift=-1ex},
                mark size=0.8pt,
            ]
                \addplot[color=orange,mark=triangle,] coordinates {
                    (2, 24.80655373267348)(3, 24.937716163113635)(4, 24.8993114916384)(5, 25.753654434704913)(6, 26.39631456765119)(7, 27.51242142376259)(8, 29.235866541476412)(16, 47.64047480907925)
                };
            \end{axis}
        \end{tikzpicture}
    \end{subfigure}
    \caption{
        Quality of images generated at framed index six, with respect to the number of simultaneously generated views.
        Additional views beyond two generated views are selected randomly in proximity of view index six.
        Beyond the number of views used during training, the quality begins to decrease as FID rises.
    }
    \label{fig:gen_sweep}
\end{figure}

Second, we more directly analyse the impact of the number of generated frames to the quality of frames for a fixed frame index in a sequence.
Given the ground-truth trajectories, we construct custom trajectories that vary the number of generated views, while conditioning on the first view in the sequence, which has a frame index of zero.
We consider a minimum number of generated views of two, where we always use the ground-truth poses at frame indices three and six.
When evaluating image quality as FID, we always measure the FID over the views at index six over all scenes.
To increase the number of generated views beyond two, additional views are given poses that are the same as the pose at index six, but with an uniform random translation with a magnitude of 10\% of the distance between the view at index six and three.
The results in \reffig{fig:gen_sweep} show that the FID remains relatively unchanged for the number of generated views up to those seen during training (i.e., four), but begins to increase noticeably beyond four generated views.

\section{Rendering Novel Views from DFM}
\label{sec:DFM}
For evaluations comparing our method with DFM \cite{tewari2023diffusion}, we obtain novel views from DFM using the publicly available pretrained weights and code.
Sampling with one target view (DFM-1) is readily supported using the available code base; however, multiple target views should be considered to ensure coverage over larger scenes.
DFM supports autoregressive generation for a variable number of target views; however, few details are available on the best methodology for rendering the intermediate non-target views in this setting.
In this section, we provide details on the generation and rendering method we use in our experiments for DFM-2 on RealEstate10K.

Given an observed image and the pose of a target view, DFM performs a generative diffusion process to generate the target view.
For multiple targets, autoregressive generation is used, where each target is diffused sequentially while conditioning on past observed or generated target views.
In our experiments with DFM-2, with two target views, we first run two iterations of autoregressive generation until all target views have been generated.
DFM creates an intermediate NeRF representation for every target view generated.
Our intermediate novel views (non-target views) are rendered from the intermediate NeRF after diffusing the final target view.

\section{Additional Results: Ground-truth Trajectories}
\label{sec:indist}
Additional qualitative results for ground-truth trajectories are provided with an interactive viewer on our \suppwebpage, for RealEstate10K and Matterport3D, under the \textbf{RealEstate10K Qualitative Results: Ground-truth Trajectories} and \textbf{Matterport3D Qualitative Results: Ground-truth Trajectories} headings.
Additional results comparing with DFM \cite{tewari2023diffusion} at a lower resolution are also provided on our \suppwebpage\ under the \textbf{DFM Qualitative Results: Ground-truth Trajectories} heading.
The viewer allows views along the trajectory to be inspected manually or with automated playback.
Different scenes along with three different samples are available.
Notice that the quality of the frames later in our keyframed generations tend to be of higher quality.

\begin{figure}[t]
    \centering
    \includegraphics[width=\linewidth]{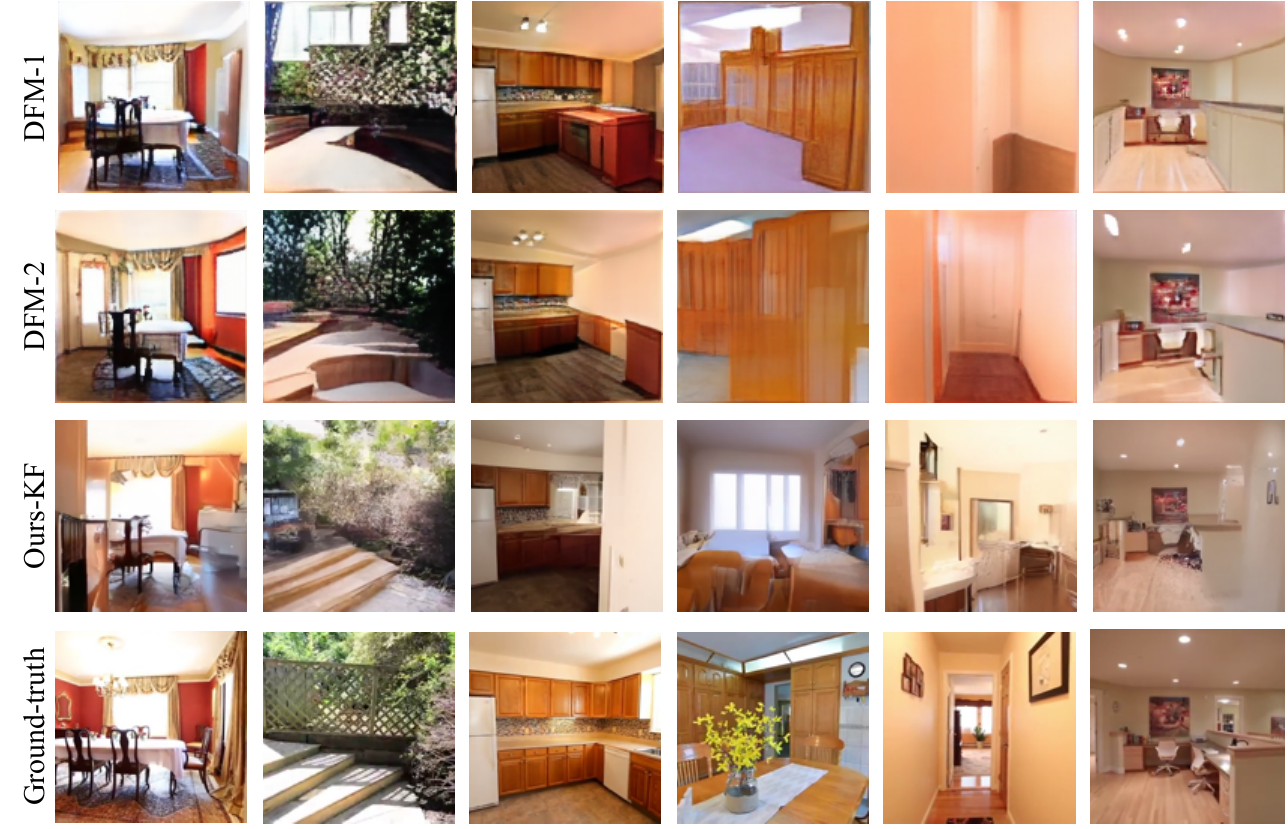}
    \caption{
        Visualization of final frames of generated sequences on ground-truth trajectories comparing our method with DFM\cite{tewari2023diffusion}.
        All images here are at the native resolution of DFM at $128\times128$.
        The quality of images from our keyframed method is higher than all other methods. 
    }
    \label{fig:goodframes-dfm}
\end{figure}

\noindent\textbf{Last frame visualization for RealEstate10K.}\ 
Additional qualitative visualizations of the last generated frames comparing our method with DFM \cite{tewari2023diffusion} at $128\times128$ are provided in \reffig{fig:goodframes-dfm}.
This visualization allows the quality of the final frames to be inspected easily.

\begin{figure}[t]
    \centering
    \includegraphics[width=\linewidth]{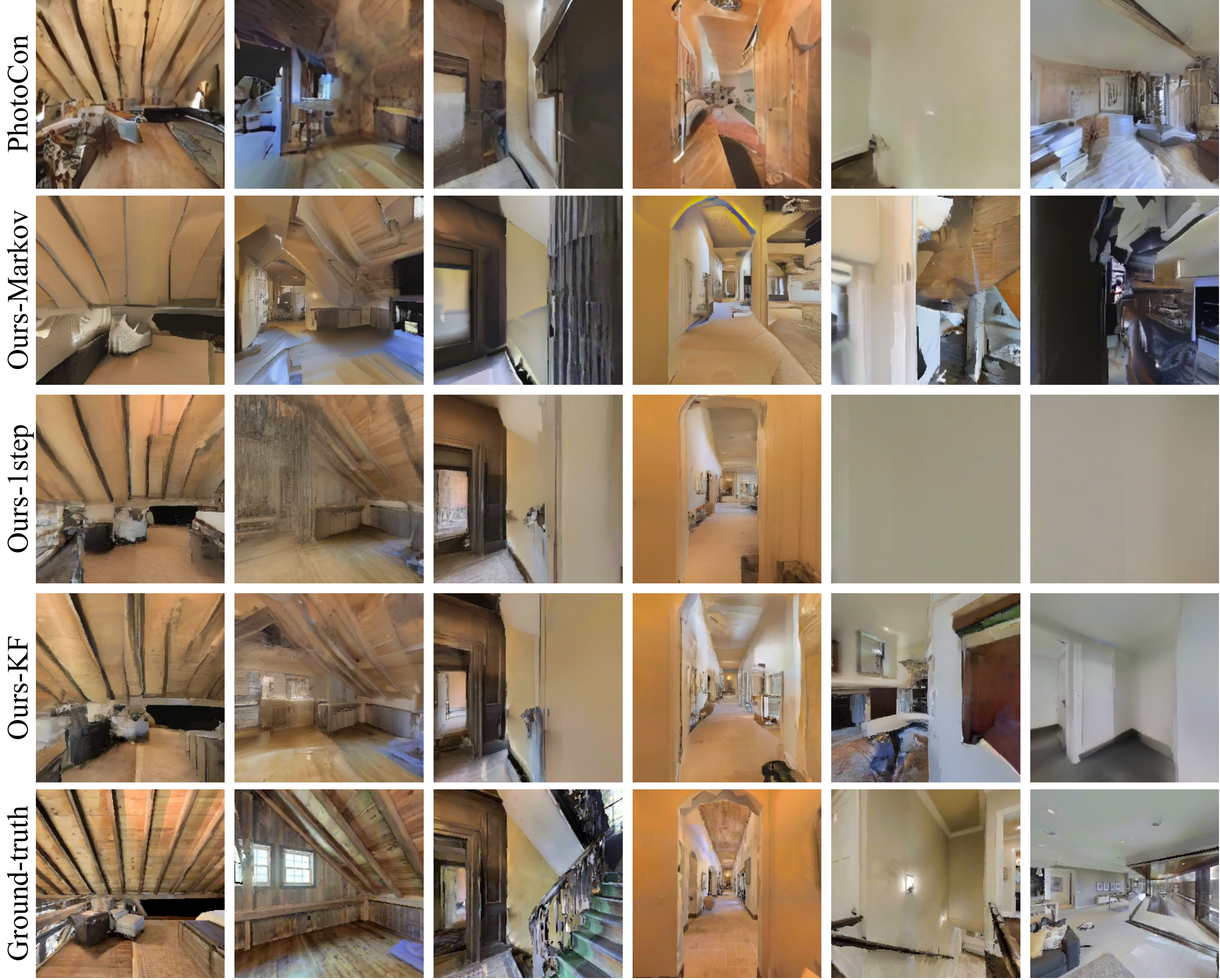}
    \caption{
        Visualization of final frames of generated sequences on ground-truth trajectories on Matterport3D.
        The quality of images from our keyframed method is higher than all other methods. 
    }
    \label{fig:goodframes-mp3d}
\end{figure}

\noindent\textbf{Last frame visualization for Matterport3D.}\ 
Additional qualitative visualizations of the last generated frames on Matterport3D are provided in \reffig{fig:goodframes-mp3d}.
Similar to the qualitative results on RealEstate10k in the main paper, the quality of the final frames on Matterport3D generated using our keyframing method are superior to the baselines.

\lccode`\1`\1
\lccode`\0`\0
\hyphenation{Real-Est-ate10k}

\section{Additional Results: Cyclical Trajectories - Spin}
\label{sec:spin}
Additional qualitative results for the \textit{spin} trajectory are provided with an interactive viewer on our \suppwebpage, under the \textbf{RealEstate10K Qualitative Results: Cyclical Trajectory - Spin} heading.
This trajectory is of particular interest due to its cyclical structure; the views eventually travel back to the original position.
Generating views with this trajectory using previous autoregressive Markov methods is challenging due to the limited conditioning window, which degrades the consistency between the last and first views, as expected.

The viewer provided on our \suppwebpage\ loops the trajectory to make the inconsistency easier to see.
There is also a \textbf{Loop First and Last} playback option to directly compare the first and last frame.
Note the scene content changes significantly in the other methods, while our keyframed method is far more consistent.

\section{Additional Results: Stereo Grouped View Generation}
\label{sec:stereo}
We provide an example stereo generation under \textbf{RealEstate10K Qualitative Results: Stereo Grouped View Generation} on our \suppwebpage.
These views contain stereo pairs along a trajectory, which do not have any natural ordering within the pairs.
A naive baseline samples these views using a standard autoregressive method, where the right then the left view is sampled before repeating for the next stereo pair in the trajectory.
Our set-based model is able to group the pairs and generate them simultaneously without imposing any ordering within the pairs.
This allows the generated stereo pairs to maintain a stable disparity along the trajectory, which is challenging using previous methods.

\section{Generating Large Sets of Unordered Views}
\label{sec:cloud}
In general, views within a set do not necessarily have a single natural ordering or any at all.
In cases where the ordering may be completely arbitrary, a heuristic based on the proximity of camera poses can be used to order the views for sampling.
Next, we describe one such simple heuristic. This is an interesting design space for future work.

Given a set of $N$ camera poses, we iteratively grow a set of keyframes, $\Omega$, as a small subset of all the frames.
Starting with the pose of the given view, $k_1$, we choose the next keyframe in $\Omega$ as a view that is not already a keyframe, and is furthest from any of the existing keyframes:
\begin{align}
    k_i = \underset{\camera_r \notin \{k_1,...k_{i-1}\}}{\arg\max}\ \underset{\camera_s \in \{k_1,...k_{i-1}\}}{\min}\ d(\camera_r,\camera_s),
    \label{eq:kf_heuristic}
\end{align}
where $k_i$ is the $i^{th}$ selected keyframe, and $d(\camera_r,\camera_s)$ quantifies a distance between the cameras.
Specifically, the distance, $d(\camera_r,\camera_s)$, is set to the Euclidean distance between the camera origins.
This heuristic is chosen to spread the keyframes out while having high coverage of the space occupied by the views.
The remaining views are used as in-between frames with a generation order defined in a similar manner to the keyframe selection method in \refeq{eq:kf_heuristic}.
The in-between frames are conditioned on a subset of the closest views that have already been generated (these may not necessarily be keyframes).

The exact algorithm used for keyframe selection is specified in \refalgo{algo:keyframe_selection}.
The generation and conditioning scheme for keyframes and in-between frames is specified in \refalgo{algo:keyframe_generation} and \refalgo{algo:between_generation}, respectively.

\begin{algorithm}[t]
    \SetAlgoLined
    Given $\textbf{C}, N_k, d(\cdot,\cdot)$: \\
    Let $\Omega \subset \mathbf{C}$ contain the camera poses of the observed views \\
    
    \SetKwFunction{FFindClosest}{FindClosest}
    \SetKwProg{Fn}{Function}{:}{end}
    \Fn{\FFindClosest{$A$,$B$}}{
        Let $a \in A$, such that $d(a,b)$ is minimized, over all $b \in B$\\
        \KwRet $a$
    }
   \ \\ 
    \SetKwFunction{FFindFurthest}{FindFurthest}
    \SetKwProg{Fn}{Function}{:}{end}
    \Fn{\FFindFurthest{$A$,$B$}}{
        Let $a \in A$, such that $d(a,$ \FFindClosest{$B,\{a\}$}$)$ is maximized\\
        \KwRet $a$
    }
   \ \\ 
    \While {$|\Omega| < N_k$}{
        Let $\textbf{C}_\text{remaining} = \textbf{C}/\Omega$ \\
        Set $\Omega \gets \Omega \cup \{$ \FFindClosest{$\textbf{C}_\text{remaining},\Omega$} $\}$
    }
    \caption{
        Keyframe selection.
        $\textbf{C}$ is the set of all camera poses.
        $N_k$ is the maximum number of keyframes.
        $d(\cdot,\cdot)$ measures the distance between camera poses.
        $\Omega$ is the set of keyframe camera poses, initialized with the observed view camera poses.
    }
    \label{algo:keyframe_selection}
\end{algorithm}

\begin{algorithm}[t]
    \SetAlgoLined
    Given $\Omega, N_\text{softlim}$: \\
    Let $\Omega_\text{avail} \subset \Omega$ contain the observed views \\
    
    \SetKwFunction{FFindClosest}{FindClosest}
    \SetKwProg{Fn}{Function}{:}{end}
    \Fn{\FFindClosest{$A$,$B$}}{
        Let $a \in A$, such that $d(a,b)$ is minimized, over all $b \in B$\\
        \KwRet $a$
    }
   \ \\ 
    \While {$\Omega \neq \Omega_\text{avail}$}{
        Set $\Omega_\text{remain} \gets \Omega/\Omega_\text{avail}$ \\
        Set $k \gets\ $\FFindClosest{$\Omega_\text{remain},\Omega_\text{avail}$}\\
        Set $c_1 \gets\ $\FFindClosest{$\Omega_\text{avail},\{k\}$}\\
        Set $c_2 \gets\ $\FFindClosest{$\Omega_\text{avail}/\{c_1\},\{k\}$}\\
        
        Set $\Omega_\text{gen} \gets \{k\}$\\
        Set $\Omega_\text{cond} \gets \{c_1,c_2\}$ \\
        \While {$ |\Omega_\text{gen}| + |\Omega_\text{cond}| < N_\text{softlim}$ \textbf{and} $\Omega \neq \Omega_\text{avail} \cup \Omega_\text{gen}$}{
            Set $k_\text{extra} \gets\ $\FFindClosest{$\Omega_\text{remain}/\Omega_\text{gen},\Omega_\text{cond}$}\\
            Set $c_\text{extra1} \gets\ $\FFindClosest{$\Omega_\text{avail},\{k_\text{extra}\}$}\\
            Set $c_\text{extra2} \gets\ $\FFindClosest{$\Omega_\text{avail}/\{c_\text{extra1}\},\{k_\text{extra}\}$}\\
            Set $\Omega_\text{gen} \gets \Omega_\text{gen} \cup \{k_\text{extra}\}$\\
            Set $\Omega_\text{cond} \gets \Omega_\text{cond} \cup \{c_\text{extra1},c_\text{extra2}\}$ \\
        }
        Sample views in $\Omega_\text{gen}$ conditioned on $\Omega_\text{cond}$\\
        Set $\Omega_\text{avail} \gets \Omega_\text{avail} \cup \Omega_\text{gen}$
    }
    \caption{
        Keyframe sampling and conditioning selection.
        Initially one keyframe is selected to be generated with two conditioning frames based on the proximity to available views (observed or already generated views).
        Additional keyframes may be selected to be simultaneously generated with this initial keyframe, which may add up to two conditioning views.
        When the total number of views ($|\Omega_\text{gen}| + |\Omega_\text{cond}|$) exceeds $N_\text{softlim}$, no additional keyframes are considered for generation.
    }
    \label{algo:keyframe_generation}
\end{algorithm}

\begin{algorithm}[t]
    \SetAlgoLined
    Given $\textbf{C}, \Omega, N_\text{cond}$: \\
    Let $\textbf{C}_\text{between} = \textbf{C}/\Omega$ \\
    Let $\textbf{C}_\text{avail} = \Omega$ \\
    
    \SetKwFunction{FFindClosest}{FindClosest}
    \SetKwProg{Fn}{Function}{:}{end}
    \Fn{\FFindClosest{$A$,$B$}}{
        Let $a \in A$, such that $d(a,b)$ is minimized, over all $b \in B$\\
        \KwRet $a$
    }
   \ \\ 
    \SetKwFunction{FFindFurthest}{FindFurthest}
    \SetKwProg{Fn}{Function}{:}{end}
    \Fn{\FFindFurthest{$A$,$B$}}{
        Let $a \in A$, such that $d(a,$ \FFindClosest{$B,\{a\}$}$)$ is maximized\\
        \KwRet $a$
    }
   \ \\ 
   \While {$\textbf{C}_\text{avail} \neq \textbf{C}$}{
        Let $c_\text{cur} =\ $\FFindFurthest{$\textbf{C}_\text{between},\textbf{C}_\text{avail}$} \\
        Set $\Omega_\text{cond} \gets \{\}$ \\
        \For {$i\gets1$ \KwTo $N_\text{cond}$} {
            Set $\Omega_\text{cond} \gets \Omega_\text{cond} \cup \{$ \FFindClosest{$\textbf{C}_\text{avail}/\Omega_\text{cond},\{c_\text{cur}\}$} $\}$ \\
        }
        Sample view $c_\text{cur}$ conditioned on $\Omega_\text{cond}$\\
        Set $\textbf{C}_\text{avail} \gets \textbf{C}_\text{avail} \cup \{c_\text{cur}\}$ \\
        Set $\textbf{C}_\text{between} \gets \textbf{C}_\text{between}/\{c_\text{cur}\}$
   }
    \caption{
        In-between frame sampling and conditioning selection.
        In-between frames are selected in a similar manner to keyframe selection, where in-between frames that are further from the generated frames are sampled first.
        Frames are generated individually while conditioning on $N_\text{cond}$ views.
    }
    \label{algo:between_generation}
\end{algorithm}

\end{document}